%% file: main.tex
\documentclass[10pt,twocolumn,letterpaper]{article}

\usepackage[pagenumbers]{cvpr} 

\newcommand\mypar[1]{\par\noindent\textbf{#1}\;\;}
\newcommand{\ourwork}{ORV\xspace}
\newcommand{\boldourwork}{\textbf{ORV}\xspace}

\usepackage[utf8]{inputenc} 
\usepackage[T1]{fontenc}    
\usepackage{fancyvrb} 

\usepackage{amsmath} 
\usepackage{multirow}
\usepackage{url}            
\usepackage[dvipsnames,table]{xcolor}         
\usepackage{xspace}
\usepackage{booktabs}       
\usepackage{amsfonts}       
\usepackage{nicefrac}       
\usepackage{microtype}      
\usepackage{colortbl}
\usepackage{enumitem}
\usepackage{algpseudocode}
\usepackage{times}
\usepackage{epsfig}
\usepackage{amssymb}
\usepackage{caption}
\usepackage{subcaption}
\usepackage{tabularx}
\usepackage{rotating}
\usepackage{etoolbox}
\usepackage[normalem]{ulem}
\usepackage{lipsum}
\usepackage{stmaryrd}
\usepackage{stackengine}
\usepackage{makecell}
\usepackage{pifont}
\usepackage{cancel}
\usepackage{wrapfig}
\usepackage{bm}
\usepackage{tablefootnote}
\usepackage{threeparttablex}
\usepackage{makecell}
\usepackage{dblfloatfix}
\usepackage{tcolorbox}
\usepackage[dvipsnames]{xcolor}
\usepackage{listings}
\tcbuselibrary{listings}
\lstdefinestyle{mystyle}{
  backgroundcolor=\color{gray!5},
  basicstyle=\ttfamily\small,
  frame=single,
  rulecolor=\color{gray!30},
  frameround=tttt,
  breaklines=true,
  captionpos=b
}

\input{preamble}
\definecolor{cvprblue}{rgb}{0.21,0.49,0.74}
\usepackage[pagebackref,breaklinks,colorlinks,allcolors=cvprblue]{hyperref}

\def\paperID{xxxxx} 
\def\confName{CVPR}
\def\confYear{2026}
\definecolor{Pad}{RGB}{30, 30, 30}
\definecolor{PadBoundary}{RGB}{255, 255, 255}
\definecolor{Act}{RGB}{237, 202, 187}
\definecolor{ActBoundary}{RGB}{255, 249, 246}

\def\model{ORV}
\title{\model: 4D \underline{O}ccupancy-centric \underline{R}obot \underline{V}ideo Generation}

\author{
Xiuyu Yang$^{1,2}$\thanks{Equal contribution} \
Bohan Li$^{3,4*}$
Shaocong Xu$^1$
Nan Wang$^1$
Chongjie Ye$^{1,5}$
Zhaoxi Chen$^{1,6}$
\\
Minghan Qin$^7$
Yikang Ding$^8$
Zheng Zhu$^9$
Xin Jin$^{4,10}$
Hang Zhao$^2$
Hao Zhao$^{1,11}$
\\[1.5mm]
\textsuperscript{1} Beijing Academy of Artificial Intelligence \quad
\textsuperscript{2} IIIS, Tsinghua University \quad
\\
\textsuperscript{3} Shanghai Jiao Tong University
\textsuperscript{4} Eastern Institute of Technology, Ningbo \quad
\\
\textsuperscript{5} The Chinese University of Hong Kong, Shenzhen
\textsuperscript{6} S-Lab, Nanyang Technological University \quad
\\
\textsuperscript{7} ByteDance \ 
\textsuperscript{8} Kuaishou \ 
\textsuperscript{9} GigaAI \
\textsuperscript{10} Zhongguancun Academy \
\textsuperscript{11} AIR, Tsinghua University
}

\begin{document}

\twocolumn[{
\renewcommand\twocolumn[1][]{#1}
\maketitle
\begin{center}
\captionsetup{type=figure}
    \vspace{-1em}
    \includegraphics[width=1.\textwidth]{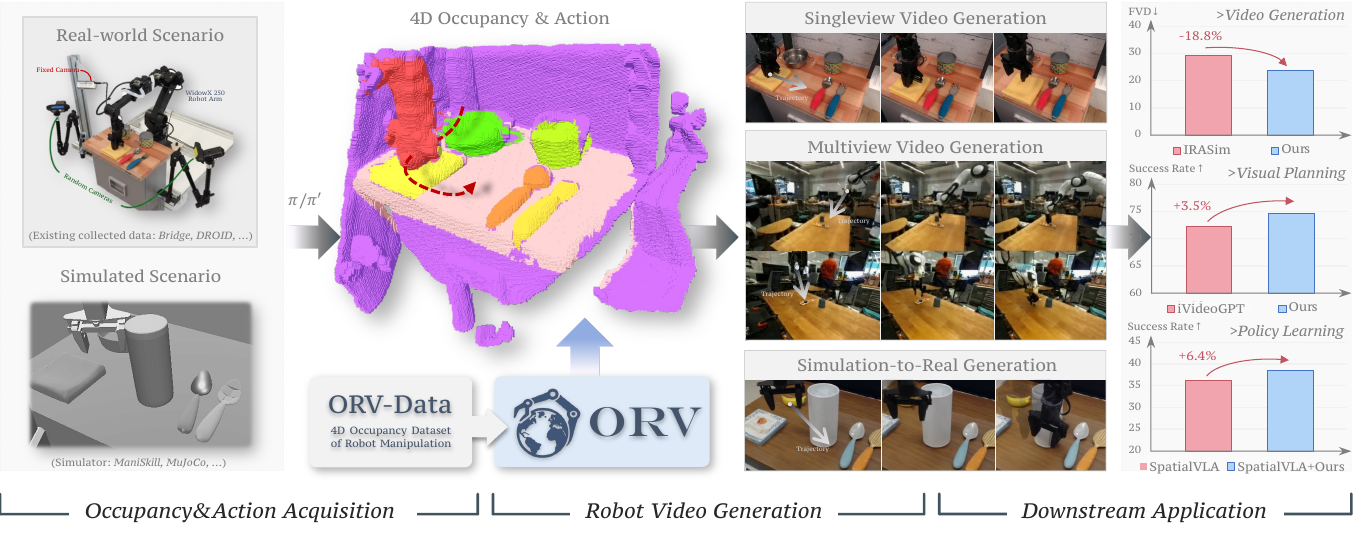}
    \vspace{-25pt}
    \captionof{figure}{
    We condition robot video generation on 4D semantic occupancy sequences and 7-DoF actions collected from real and simulated environments (through methods $\pi$ and $\pi^\prime$). This occupancy-centric conditioning enables faithful, controllable synthesis of single-view, multi-view, and sim-to-real manipulation videos. We also introduce \ourwork-Data, a curated 4D occupancy dataset for robot manipulation. Across benchmarks and downstream tasks, \ourwork improves video quality and control alignment, boosting visual planning and policy learning.
    }
    \label{fig:teaser}
\end{center}
}]

\input{sections_cvpr/0_abstract}    
\input{sections_cvpr/1_introduction}

\input{sections_cvpr/2_related_work}
\input{sections_cvpr/3_method}
\input{sections_cvpr/4_experiment}
\input{sections_cvpr/5_conclusion}
{
    \small
    \bibliographystyle{ieeenat_fullname}
    \bibliography{main}
}

\input{sections_cvpr/X_suppl}

\end{document}

%% file: sections_cvpr/0_abstract.tex
\begin{abstract}

\vspace{-3pt}
Recent embodied intelligence suffers from data scarcity, while conventional simulators lack visual realism. Controllable video generation is emerging as a promising data engine, yet current action-conditioned methods still fall short: generated videos are limited in fidelity and temporal consistency, poorly aligned with controls, and often constrained to singleview settings. We attribute these issues to the representational gap between sparse control inputs and dense pixel outputs. Thus, we introduce ORV, a 4D occupancy-centric framework for robot video generation that couples action priors with occupancy-derived visual priors. Concretely, we align chunked 7-DoF actions with video latents via an Action-Expert AdaLN modulation, and inject 2D renderings of 4D semantic occupancy into the generation process as soft guidance. Meanwhile, a central obstacle is the lack of occupancy data for embodied scenarios; we therefore curate ORV-Data, a large-scale, high-quality 4D semantic occupancy dataset of robot manipulation. Across BridgeV2, DROID, and RT-1, ORV improves video generation quality and controllability, achieving 18.8\% lower FVD than state of the art, +3.5\% success rate on visual planning, and +6.4\% success rate on policy learning. Beyond singleview generation, ORV natively supports multiview consistent synthesis and enables simulation-to-real transfer despite significant domain gaps.
Code, models, and data will be released upon acceptance.\looseness=-1

\vspace{-10pt}

\end{abstract}

%% file: sections_cvpr/1_introduction.tex
\section{Introduction}

Developing realistic simulators for robot manipulation is crucial for scaling embodied learning~\cite{katara2024gen2sim,mandi2022cacti,lee2023scale,lin2024data}. While existing simulators ~\cite{gu2023maniskill2,todorov2012mujoco} enable safe policy training and efficient data collection, they often struggle to deliver visual realism. Recent progress in generative world models ~\cite{kong2025hunyuanvideosystematicframeworklarge,wan2025wanopenadvancedlargescale,yang2024cogvideox}, especially action-conditioned video generation, offers a promising alternative by simulating future visual states conditioned on agent actions. These models can render realistic RGB observations responsive to control inputs, yet they still fall short of serving as reliable simulators: generated sequences frequently lack temporal consistency, action alignment, and multiview coherence. Bridging this gap between sparse robot controls and dense visual dynamics remains an open challenge toward building truly \textit{high-fidelity}, \textit{versatile}, and \textit{reliable} generative simulators.

Previous works~\cite{agarwal2025cosmos,zhu2024irasim,wang2025sampo,wang2025hma,rigter2024avid} have advanced action-conditioned video generation using diffusion-based or autoregressive backbones, where robot actions are typically represented as 7-DoF end-effector (EE) poses that guide visual rollout. Other studies~\cite{yang2023learning,huang2025enerverse,zhen2025tesseract} instead employ high-level conditioning such as language instructions to drive scene dynamics. Despite these advances, existing approaches remain constrained by three key limitations:
(\texttt{p1}) limited visual fidelity and temporal consistency;
(\texttt{p2}) drifted or misaligned future predictions that fail to reflect manipulation controls faithfully; and
(\texttt{p3}) restriction to singleview observations without enforcing multiview coherence.

\begin{figure}
  \centering
  \includegraphics[width=\linewidth]{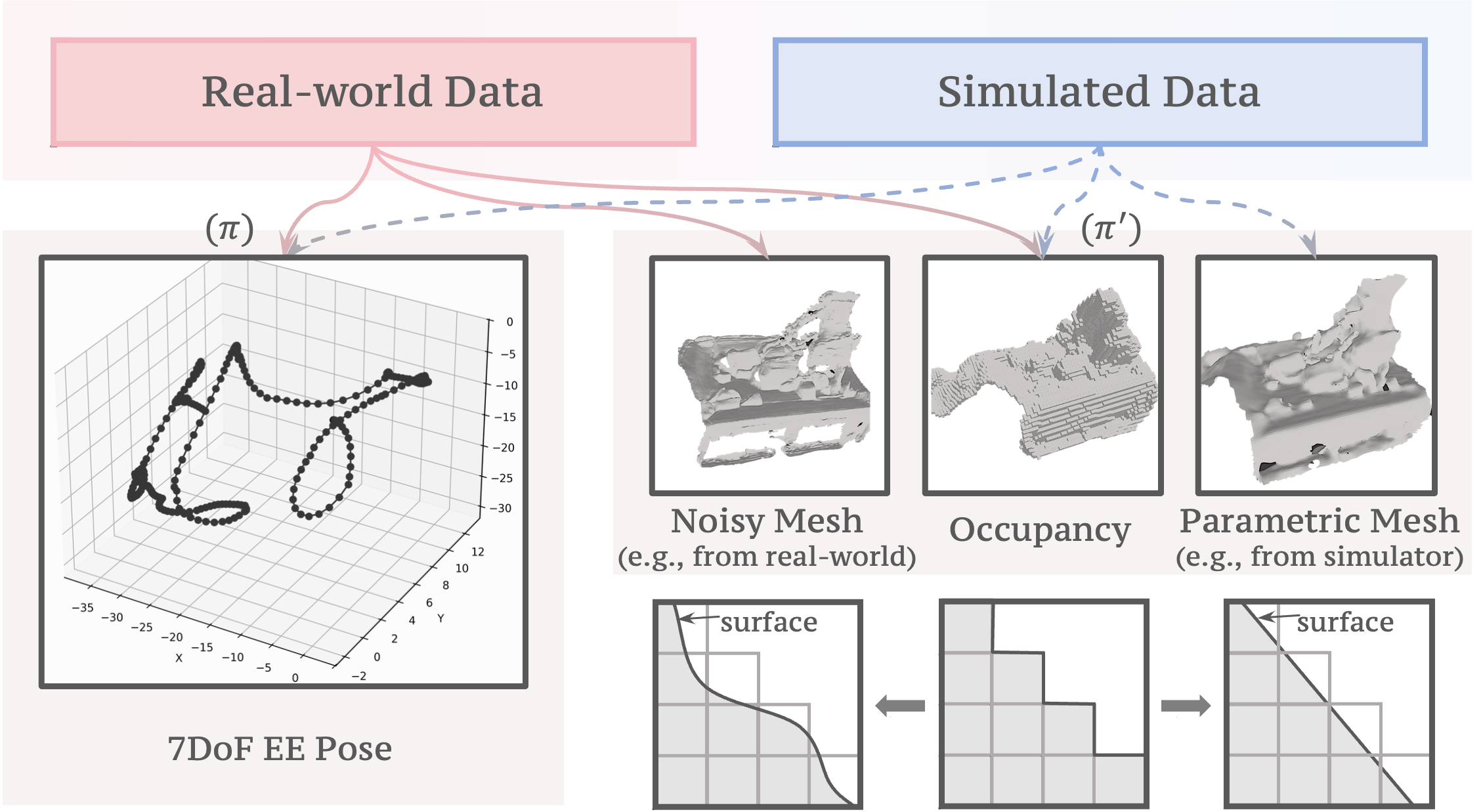}
  \captionsetup{font=footnotesize}
  \caption{Establishment of non-interactive methods $(\pi,\pi^{\prime})$ both in the real-world environment and physical simulator to collect trajectory priors (7-DoF EE Pose) and visual priors (Occupancy).}
  \label{fig:intro_occ}
  \vspace{-15pt}
\end{figure}

We propose \ourwork, a versatile 4D occupancy-centric framework for robot video generation that produces high-fidelity, action-aligned visual simulations. Our key insight is to incorporate 4D semantic occupancy as visual priors that complement conventional action priors, effectively bridging the representational gap between sparse control trajectories and dense visual dynamics. We think that limitations \texttt{p2}, \texttt{p3} largely stem from this gap, as also observed in prior works~\cite{wang2025precise,li2025mask2ivinteractioncentricvideogeneration,xu2024flow,liu2025robotransfer} which introduce fine-grained cues such as optical flow, masks, or skeletons to enhance controllability. Furthermore, as illustrated in Fig.~\ref{fig:intro_occ}, occupancy fields demonstrate robustness to geometric noise, providing a natural bridge between simulated and real-world scenarios. Moreover, \ourwork leverages the generative capabilities of modern video foundation models~\cite{yang2024cogvideox,kong2025hunyuanvideosystematicframeworklarge,wan2025wanopenadvancedlargescale} to boost visual realism and temporal coherence, substantially mitigating issue (\texttt{p1}) while preserving physically consistent dynamics.

The overall framework of \ourwork is depicted in Fig.~\ref{fig:teaser}. Guided by geometric priors from 4D semantic occupancy, \ourwork enables robot manipulation video generation across diverse object appearances and scenes~\cite{liu2025robotransfer,alhaija2025cosmos}. Furthermore, view-specific conditioning encourages cross-view coherence, enabling consistent multiview synthesis~\cite{acar2023visual,goyal2023rvt,asali2023mvsa}. Benefiting from the domain-invariant nature of occupancy-derived representations, \ourwork also facilitates visual transfer from simulation to the real world under varied conditions. To support large-scale training, we curate ORV-Data, a high-quality 4D semantic occupancy dataset for robot manipulation, built through a carefully designed data curation pipeline.

Our contributions can be summarized as follows:
\begin{itemize}
    \item We propose \textbf{\ourwork}, a \textit{4D occupancy-centric framework}, enabling precise and controllable robot video generation with domain randomization.

    \item By injecting \textit{occupancy-derived geometric priors} into diffusion noise, \ourwork achieves temporally consistent and geometrically coherent multiview video generation and simulation-to-real visual transfer.
    \item We curate \textbf{\ourwork-Data}, a large-scale, high-quality \textit{4D semantic occupancy dataset} of robot manipulation with rich geometric and semantic annotations.
    \item Experiments across diverse datasets and downstream tasks demonstrate that \ourwork consistently enhances controllable video generation, visual planning, and data-driven policy learning, achieving state-of-the-art performance.
\end{itemize}

%% file: sections_cvpr/2_related_work.tex
\begin{figure*}
  \centering
  \includegraphics[width=\linewidth]{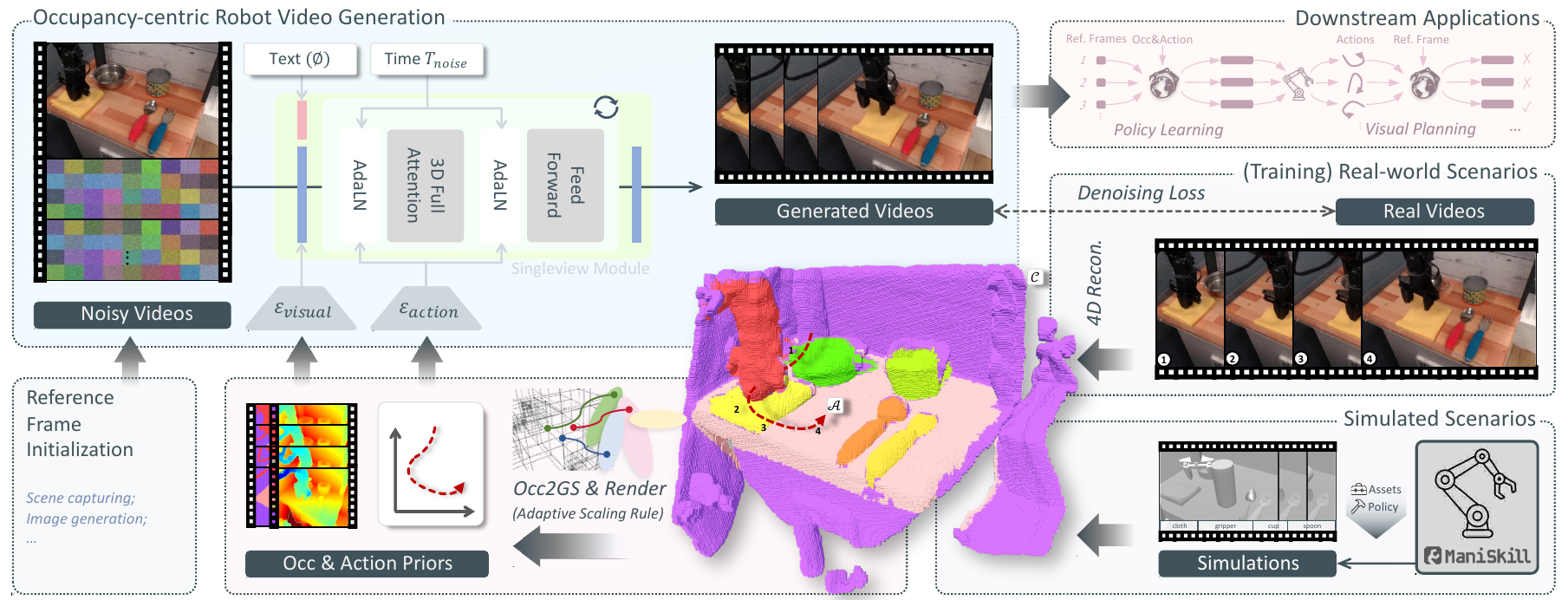}
  \caption{Overview of \boldourwork \textbf{framework}. Centered on occupancy representation $\mathcal{C}$, along with actions $\mathcal{A}$, which are extracted from physical simulators (\textit{e.g.}, ManiSkill~\cite{gu2023maniskill2}) or real-world data (\textit{e.g.}, Bridge~\cite{walke2023bridgedata}), we leverage the soft visual priors to enable robot video generation with high visual quality and control alignment. Furthermore, we design a data curation pipeline to construct the robot occupancy data for training purposes. ORV, as a powerful neural simulator, can greatly boost downstream applications (\textit{e.g.}, policy learning, visual planning, etc).\looseness=-1}
  \label{fig:sys}
  \vspace{-13pt}
\end{figure*}

\section{Related Work}

\mypar{Generative Models for World Modeling.}
Recent advances in video generation~\cite{zheng2024opensora,yang2024cogvideox,wan2025wanopenadvancedlargescale,kong2025hunyuanvideosystematicframeworklarge,bai2025recammaster,bai2024syncammaster,yu2025trajectorycrafter} have greatly improved the realism of world modeling, benefiting robotics~\cite{zhu2024irasim,wang2025precise,zhen2025tesseract,qian2025wristworld,liu2025robotransfer,fu2025learningvideogenerationrobotic,luo2024grounding,jiang2024dexmimicgen,bharadhwaj2024gen2act,alhaija2025cosmos}, autonomous driving~\cite{gao2023magicdrive,li2024uniscene,wang2024driving,mao2024dreamdrive}, and general scene synthesis~\cite{lin2025controllable,ren2025gen3c,zhangjie2025difix3d+,liang2025diffusionrenderer}.
ReCamMaster~\cite{bai2025recammaster} and SynCamMaster~\cite{bai2024syncammaster} achieve video synthesis of novel trajectories, while IRASim~\cite{zhu2024irasim} enables action-to-video prediction, and VAP~\cite{wang2025precise} employs visual prompts for precise control in robotics.
For autonomous driving, more recent works adopt 3D occupancy as efficient scene representations~\cite{li2024hierarchical,li2023bridging,cao2022monoscene,huang2023tri,wei2023surroundocc,wang2024occsora,zheng2023occworld,wei2024occllama,huang2024gaussianformer,wu2024embodiedocc,wang2024embodiedscan}. For instance, UniScene~\cite{li2024uniscene} leverages hierarchical occupancy priors for multimodal scene generation.
Beyond explicit video synthesis, implicit generative models have also been adopted for complex interactions and decision making~\cite{netanyahu2024few,lyu2025dywa,cherniavskiistream}.

\mypar{World Models for Embodied Intelligence.}
Progress in simulating dynamic environments has fueled the development of world models for robotics~\cite{yang2023learning,huang2025enerverse,bharadhwaj2024gen2act,cheang2024gr,geng2025roboverse,chi2024eva,bruce2024genie,zhou2024robodreamer,zhu2025unified,zhang2024combo,bjorck2025gr00t,guo2025ctrl,long2025learning},
where TesserAct~\cite{zhen2025tesseract} performs 4D scene synthesis via appearance--geometry joint modeling and EnerVerse~\cite{huang2025enerverse} forecasts future environments through a simulation pipeline.
iVideoGPT~\cite{wu2024ivideogpt} and Vid2World~\cite{huang2025vid2world} explore action-conditioned visual prediction with autoregressive frameworks.
For data augmentation, Cosmos-Transfer~\cite{alhaija2025cosmos} and RoboTransfer~\cite{liu2025robotransfer} condition robot video generation on scene maps (\textit{e.g.}, depth and normal), while RoboEngine~\cite{yuan2025roboengine} achieves scene augmentation through the segmentation toolkit. 
Meanwhile, WorldSimBench~\cite{qin2024worldsimbench} establishes unified evaluation benchmarks for world models.

%% file: sections_cvpr/3_method.tex
\section{\ourwork: Methodology}

We first formulate the robot video generation task (Sec.~\ref{sec:formulation}). Then we elaborate on the specific architecture of \ourwork and how these designs can largely improve the robot video generation (Sec.~\ref{sec:vgm}). Finally, we introduce our robot occupancy dataset curated for the training process (Sec.~\ref{sec:data}) and explain how \ourwork helps with the robot manipulations.

\vspace{-3pt}
\subsection{Problem Formulation}
\label{sec:formulation}

A generative world model for robot manipulation aims to provide a photorealistic and physically consistent simulation of the environment that mirrors real-world dynamics.
Given the context $(\mathcal{S}, \mathcal{O}, \phi, \rho)$, the goal of the model $\mathcal{M}$ is to predict future states $s_{t:t+\Delta T}\!\in\!\mathcal{S}$ and corresponding observations $o_{t:t+\Delta t}\!\in\!\mathcal{O}$, where $o_t=\phi(s_t)$ denotes the rendered observation from state $s_t$.
Here, $\rho$ defines the underlying rules governing state transitions, leading to the transition probability $p(s_{t:t+\Delta t}, o_{t:t+\Delta t}\mid s_{1:t}, o_{1:t})$.

We formulate $\mathcal{O}$ in RGB space (\textit{e.g.}, images or videos). Conventional text-to-video models~\cite{yang2024cogvideox,wan2025wanopenadvancedlargescale,li2024hunyuan} condition on $\rho_1\!:=\!\text{{\fontfamily{pcr}\selectfont Embed}}(\text{{\fontfamily{pcr}\selectfont text}})$, yet linguistic abstraction often hinders accurate physical simulation.
Recent action-conditioned video generation~\cite{wang2025precise,wu2024ivideogpt,wang2025hma,rigter2024avid} extends this to $\rho_2\!:=\!\text{{\fontfamily{pcr}\selectfont Embed}}(a_{t:t+\Delta t}\!\sim\!\pi(s_{1:t}))$.
Building upon this progression, our model introduces $\rho_3\!:=\!\text{{\fontfamily{pcr}\selectfont Embed}}(c_{t:t+\Delta t}\!\sim\!\pi'(s_{1:t}), a_{t:t+\Delta t}\!\sim\!\pi(s_{1:t}))$, where $a$ denotes agent actions and $c$ represents occupancy fields.
We denote by $\pi$ and $\pi'$ the extraction processes for ($a,c$) given states $s$.

As illustrated in Fig.~\ref{fig:intro_occ}, both extraction methods can be established either in the real world (\textit{e.g.}, human teleoperation) or within simulators (\textit{e.g.}, ManiSkill~\cite{gu2023maniskill2}, MuJoCo~\cite{todorov2012mujoco}).
Notably, we employ $\pi$ and $\pi^{\prime}$ in a \textit{non-interactive} manner---these priors are collected entirely in a single offline pass before being used.
Moreover, the motivation for leveraging occupancies lies in their robustness for representing both noisy and parametric scene surfaces (Fig.~\ref{fig:intro_occ}).
And the coordinate-based formulation of occupancies enables seamless integration with online occupancy generations~\cite{zhang2025roboocc}.

\vspace{-3pt}
\subsection{Occupancy-centric Robot Video Generation}
\label{sec:vgm}
To avoid a costly large-scale pretraining process (as previous works~\cite{wu2024ivideogpt,zhu2025unified}) and reduce the training cost, we build \ourwork model upon the pretrained open-source models (\textit{e.g.}, we use CogVideoX-2B~\cite{yang2024cogvideox}), which also aligns with our non-interactive purpose (using a bidirectional diffusion model).
CogVideoX incorporates the architecture of diffusion transformer (DiT) and achieves incredible performance.
Then, we propose a two-stage supervised finetuning (SFT) to inject both action and visual cues into video generations.
We aim to address three key aspects: 1) overall quality of generated videos (\textit{e.g.}, consistency of frames and realism), 2) alignment with the instructions $\rho_3$, and 3) computation efficiency.\looseness=-1

\begin{figure}
  \centering
  \vspace{-2mm}
  \includegraphics[width=\linewidth]{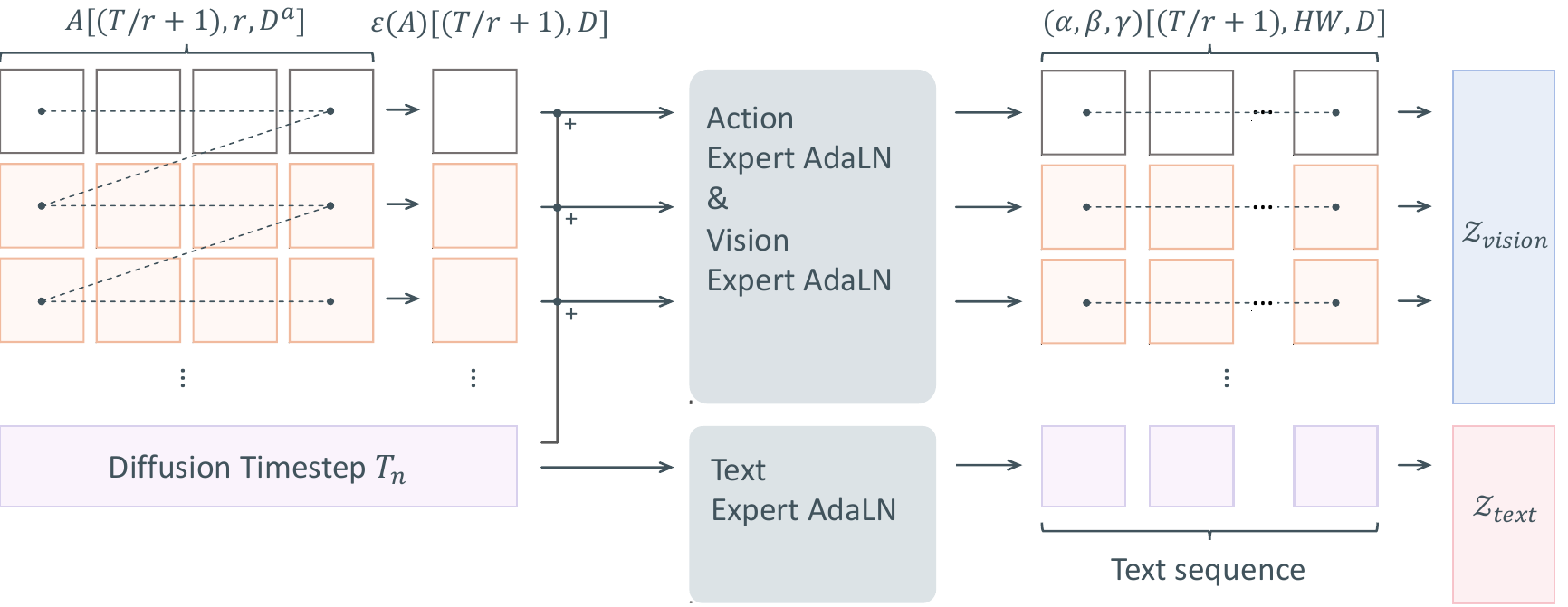}
  \captionsetup{font=footnotesize}
  \caption{Illustration of three modulations (Expert AdaLN) and injecting actions \raisebox{0.4ex}{\fcolorbox{Act}{ActBoundary}{\makebox[1pt]{\phantom{\rule{0pt}{1pt}}}}} in our DiT block. And \raisebox{0.4ex}{\fcolorbox{Pad}{PadBoundary}{\makebox[1pt]{\phantom{\rule{0pt}{1pt}}}}} indicates the action paddings serving as the placeholders for reference frames, where $\varepsilon$ encodes actions and $\alpha,\beta,\gamma$ are modulation vectors. We use $[\cdot]$ to indicate the dimensions for simplicity.}
  \label{fig:mod}
  \vspace{-5pt}
\end{figure}

\mypar{Chunk-level Action Conditioning.}
The 7-DoF action sequences (\textit{e.g.}, $\mathcal{A}\!\in\!R^{T\times D_{a}}$ derived from end-effector pose sequences and $D_a\!=\!7$) serve a high-level control signals in robot video generation.
Drawing inspiration from~\cite{zhu2024irasim,zhang2024tora}, we inject these 3D action controls through adaptive layer normalization (Action Expert AdaLN) to directly modulate the video latents within each DiT block.
More efficiently, as illustrated in Fig.~\ref{fig:mod}, we propose a chunk-level scheme for temporal alignment between high-dimensional actions and videos in modulation.\looseness=-1

Specifically, following the temporal compression in 3D VAE~\cite{yang2024cogvideox,kong2025hunyuanvideosystematicframeworklarge,wan2025wanopenadvancedlargescale}, we pad zero actions as the placeholders of reference frames.
Then an additional shallow MLP ($\varepsilon_{action}$ in Fig.~\ref{fig:sys}) is used to map every consecutive $r$ actions into a single token: $\mathcal{A}\in R^{T\times D_a}\rightarrow \text{{\fontfamily{pcr}\selectfont MLP}}(\text{{\fontfamily{pcr}\selectfont Pad}}(\mathcal{A}))\in R^{(\frac{T}{r}+1)\times D}$, where $r$ denotes the chunk-size and $D$ represents the feature size.
Furthermore, we let Action Expert AdaLN reuse the parameters of pretrained Vision Expert AdaLN, eliminating the unnecessary computation cost (as each AdaLN accounts for $\sim1/3$ of the total parameters).

\mypar{Occupancy-derived Visual Conditioning.}
Translating abstract 3D action signals into 2D pixels presents a great challenge; thus, we introduce \textit{soft} and \textit{pixel-level} visual conditionings derived from occupancy fields.
However, directly projecting voxels onto 2D planes will cause mutations on pixels between adjacent frames and viewpoints.
We further propose to assign each grid with non-learnable Gaussian splatting~\cite{kerbl20233d}, then render them from certain views (Fig.~\ref{fig:sys}), which greatly improves the conditions quality and saves memory.\looseness=-1

Moreover, we propose an \textit{adaptive scaling mechanism} on Gaussians to solve the perspective distortion during rendering (see Sec.~\ref{sec:supp_data_render} in Suppl. for derivations).
Specifically, the scale follows $\sigma=k_2\cdot \hat{z}^{k_1}$, where $\hat{z}\in[1, 2)$ denotes the \textit{normalized depths in canonical space}, and exponential term $k_1$, base scale term $k_2$ control the scaling behavior of Gaussians in the near and far plane, respectively.

To inject such occupancy-derived visual conditionings, we deploy an additional encoder MLP ($\varepsilon_{visual}$ in Fig.~\ref{fig:sys}), then augment it with the input images, after which another zero-initialized projector adds the visual conditionings to the input noise: $z_{\text{in}}=\text{{\fontfamily{pcr}\selectfont Zero-MLP}}(z_{\text{in}}+\text{{\fontfamily{pcr}\selectfont MLP}}(\mathcal{C})) + z_{\text{in}}$.
The previous ControlNet-like~\cite{zhang2023adding} methods, though demonstrating accurate controls, suffer from a serious computation cost (see Sec.~\ref{sec:supp_model_arch} in Suppl.).
Furthermore, such layer-wise control injection tends to corrupt the video latents when conditions are \textit{soft}---that is, not pixel-level alignment with ground truth.

\subsubsection{\ourwork-MV: Multiview Robot Video Generation}
\label{sec:mv}

A complete and high-fidelity 4D world, typically formed from multiview observations, greatly benefits robot learnings~\cite{liu2025geometry,qian2025wristworld}.
Leveraging the 4D occupancy-centric design, ORV(-MV) generates multiview robot manipulation videos well.
Some prior works~\cite{zhen2025tesseract,zhu2024irasim,wang2025precise}, however, capture only a single surface of the scenes, resulting in noticeable artifacts and empty regions in shifted views.

\begin{figure}
  \centering
  \includegraphics[width=\linewidth]{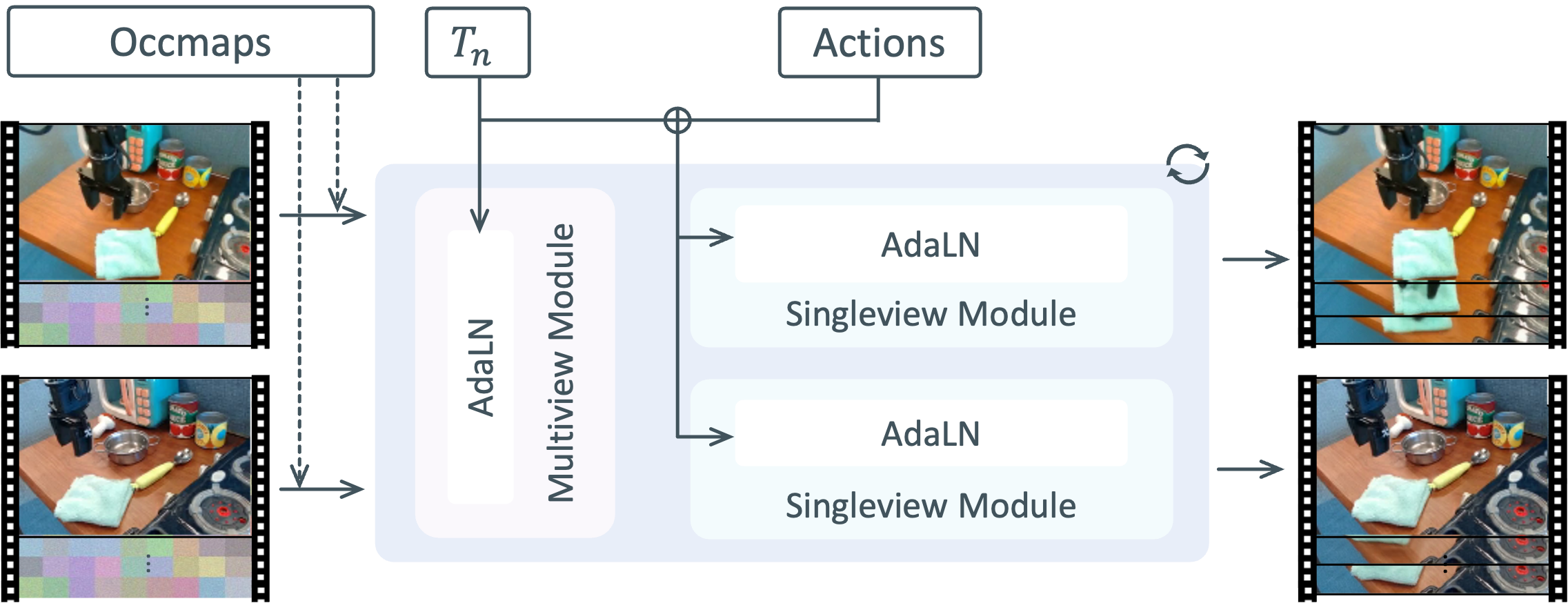}
  \captionsetup{font=footnotesize}
  \caption{Architecture of \ourwork-MV, which generates multiview robot manipulation videos with cross-view consistency.}
  \label{fig:mv_pipeline}
  \vspace{-15pt}
\end{figure}

As shown in Fig.~\ref{fig:mv_pipeline}, \ourwork-MV introduces an additional view attention (multiview module) prior to the temporal attention (singleview module), inspired by~\cite{bai2024syncammaster,cao2024mvgenmaster}.
Both inherit the 3D (2D+1D) attention layers of the pretrained model, with 2D over pixels $H\!\times\!W$ and 1D over views $V$ or frames $F$.
The former processes the latents $\mathcal{F}_V\!\in\!R^{B_V\times S_V\times D}$, where $S_V\!=\!VHW$ denotes patch tokens across all views.
While the latter handles $\mathcal{F}_P\!\in\!R^{B_P\times S_P\times D}$, where $S_P\!=\!THW$ denotes tokens across all times of each view.

We then apply different controls for the two modules.
Specifically, singleview modules are conditioned on text, actions, and occmaps.
While multiview ones exclude action priors, as they focus on view correspondences.
Additionally, details on handling multiview occupancy map data for training purposes are provided in Sec.~\ref{sec:supp_mv} in Suppl.

\begin{figure*}
    \centering
    \includegraphics[width=1.\textwidth]{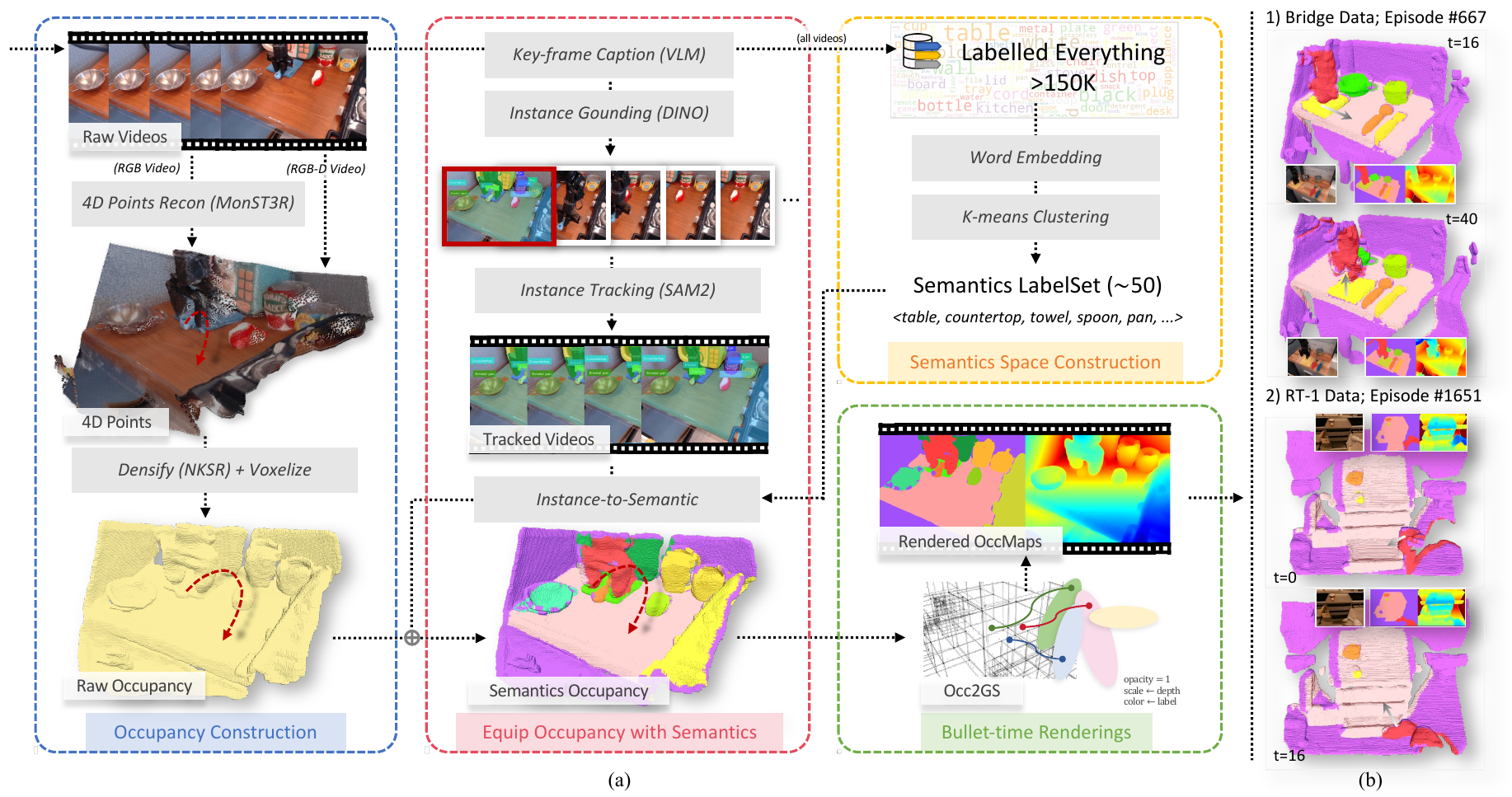}
    \vspace{-5pt}
    \caption{(a) Overview of \textbf{Training Dataset Curation Pipeline}, which consists of four steps: 1) semantics space construction, 2) occupancy construction, 3) equip occupancy with semantics, and 4) bullet-time occupancy-to-Gaussian renderings in practical usage. (b) \textbf{Occupancy examples} of BridgeData V2 (BridgeV2)~\cite{walke2023bridgedata} and RT-1~\cite{brohan2022rt1}. Better to zoom in. Refer to Supplementary Materials for more examples.}
    \label{fig:data}
    \vspace{-12pt}
\end{figure*}

\subsubsection{\ourwork-S2R: Bridge Simulation-to-Real Transfer}
\label{sec:sim2real}
The occupancy-derived visual priors (\textit{e.g.}, depth maps) also enable ORV(-S2R) to generate realistic videos from such appearance-agnostic information, which is crucial for alleviating the \textit{visual realism} gap between simulated and real data in robotics.
As shown in Fig.~\ref{fig:sys}, physical simulators (\textit{e.g.}, ManiSkill~\cite{gu2023maniskill2}, MuJoCo~\cite{todorov2012mujoco}) can readily provide such priors at a low cost.

Previous works, \textit{e.g.}, Cosmos-Transfer~\cite{alhaija2025cosmos}, RoboTransfer~\cite{liu2025robotransfer}, have also demonstrated success in transferring multi-modal data to significantly mitigate the data scarcity problem in robotics.
However, as described in Sec.~\ref{sec:formulation} and Fig.~\ref{fig:intro_occ}, the occupancy-derived condition maps further exhibit robustness to geometric noise, providing a natural bridge between real-world noisy surfaces and parametric ones in simulation.
Experiments in Sec.~\ref{sec:exp_abla} have validated the effectiveness of this design.\looseness=-1

\subsection{4D Occupancy Dataset of Robot Manipulation}
\label{sec:data}
To train \ourwork model, we establish a 4D occupancy dataset of robot manipulation through the data curation pipeline shown in Fig.~\ref{fig:data}(a).
The occupancy data are derived from existing popular robot datasets (BridgeData V2~\cite{walke2023bridgedata}, DROID~\cite{khazatsky2024droid}, RT-1~\cite{brohan2022rt1}). Some examples are shown in Fig.~\ref{fig:data}(b).
More details are provided in Sec.~\ref{sec:supp_occ_data} and Sec.~\ref{sec:more_occ} in Suppl.

\mypar{Semantics Labeling.}
Complex semantic understanding remains essential in robot manipulation, as predicting next-state dynamics requires recognizing objects—rigid, articulated, or deformable—that exhibit distinct physical behaviors.
To this end, we construct the dataset-level semantic space through vision-language-model (VLM)~\cite{bai2023qwen} for captioning and K-means~\cite{lloyd1982least} clustering over $\sim$150K labels.
For each video, we then extract temporally consistent instances across frames using Grounding DINO~\cite{liu2024grounding} and SAM2~\cite{ravi2024sam}, after which they are mapped to coherent semantics.

\mypar{4D Occupancy Generation.}
This process involves two steps: 1) occupancy construction and 2) semantic enrichment.
We first reconstruct sparse 4D points with MonST3R~\cite{zhang2024monst3r} and then densify them by NKSR~\cite{huang2023nksr}, which greatly fills holes and is robust to noise.
Note that for those videos with a depth channel, the reconstruction is not needed.
Then, dense points are voxelized to 4D occupancy in canonical space, after which semantics are assigned by majority voting for points with projected semantic labels within each voxel.
Finally, we filter the occupancy-rendered data with poor inter-frame consistency (through RAFT~\cite{teed2020raft}).

\renewcommand{\arraystretch}{0.9}
\begin{table*}[t]
    \centering
    \small
    \caption{Evaluation results of \textit{Conditional Video Generation} on three datasets. Top-1 performance within all variants and each type of model is represented with \textbf{bold text} and \underline{underlines}.}
    \vspace{-2pt}
    \renewcommand{\arraystretch}{1.0}
    \setlength{\tabcolsep}{3.5pt}
    \begin{threeparttable}
    \begin{tabularx}{0.99\linewidth}{l *{12}{>{\centering\arraybackslash}X}}
        \toprule[1pt]
        \multirow{2}{*}{Method}
        & \multicolumn{4}{c}{\cellcolor{green!0}BridgeData V2~\cite{walke2023bridgedata}} 
        & \multicolumn{4}{c}{\cellcolor{blue!0}DROID~\cite{khazatsky2024droid}} 
        & \multicolumn{4}{c}{\cellcolor{brown!0}RT-1~\cite{brohan2022rt1}} \\
        \cmidrule(lr){2-5} \cmidrule(lr){6-9} \cmidrule(lr){10-13}
        & PSNR$\uparrow$ & SSIM$\uparrow$ & FID$\downarrow$ & FVD$\downarrow$ 
        & PSNR$\uparrow$ & SSIM$\uparrow$ & FID$\downarrow$ & FVD$\downarrow$
        & PSNR$\uparrow$ & SSIM$\uparrow$ & FID$\downarrow$ & FVD$\downarrow$ \\
        \midrule[1pt]
        \rowcolor{cyan!5}
        \multicolumn{13}{c}{\textit{Text-conditioned Generation Models}} \\
        CogVideoX~\cite{yang2024cogvideox} & 19.432 & 0.752 & 7.509 & 83.561 & 19.238 & 0.701 & 6.341 & 71.536 & 20.457 & 0.816 & 6.243 & 42.169 \\
        \midrule[0.5pt]
        \rowcolor{red!5}
        \multicolumn{13}{c}{\textit{Action-conditioned Generation Models}} \\
        AVID~\cite{rigter2024avid} & - & - & - & - & - & - & - & - & 25.600 & 0.852 & \textcolor{gray}{2.965}\rlap{\raisebox{2pt}{\tnote{*}}} & 24.200 \\
        HMA~\cite{wang2025hma} & 23.636 & 0.808 & 8.849 & 67.096 & 21.435 & 0.821 & \underline{\textbf{3.108}} & 47.383 & 25.424 & 0.840 & 7.306 & 84.165 \\
        IRASim~\cite{zhu2024irasim} & 25.276 & 0.833 & 10.510 & 20.910 & 21.632 & 0.820 & 5.395 & 41.031 & 26.048 & 0.833 & 5.600 & 25.580 \\
        \ourwork (Ours) & \underline{25.631} & \underline{0.873} & \underline{3.821} & \underline{17.682} & \underline{22.034} & \underline{0.838} & 4.921 & \underline{37.094} & \underline{27.086} & \underline{0.863} & \underline{4.210} & \underline{20.031} \\
        \midrule[0.5pt]
        \rowcolor{green!5}
        \multicolumn{13}{c}{\textit{Occupancy\&Action-conditioned Generation Models}} \\
        IRASim$^{\dagger}$~\cite{zhu2024irasim} & 27.352 & 0.862 & 9.413 & 22.503 & 22.005 & 0.827 & 7.892 & 44.309 & 27.213 & 0.847 & 5.311 & 42.130 \\
        \ourwork (Ours) & \underline{\textbf{28.258}} & \underline{\textbf{0.899}} & \underline{\textbf{3.418}} & \underline{\textbf{16.525}} & \underline{\textbf{22.310}} & \underline{\textbf{0.841}} & \underline{3.222} & \underline{\textbf{34.603}} & \underline{\textbf{28.214}} & \underline{\textbf{0.878}} & \underline{\textbf{4.013}} & \underline{\textbf{19.931}} \\
        \bottomrule[1pt]
    \end{tabularx}

    \begin{tablenotes}
    \footnotesize
    \item[*] FID Scores of AVID~\cite{rigter2024avid} have been computed not in evaluation mode according to the \href{https://github.com/microsoft/causica/blob/main/research_experiments/avid/libs/avid_utils/avid_utils/metrics.py}{official codes} and lead to incorrect results. Thus, we ignore it.
    \item[$\dagger$] We incorporate the same occupancy\&action conditions to IRASim.
    \end{tablenotes}

    \end{threeparttable}
    \label{tab:video_gen}
    \vspace{-1pt}
\end{table*}

\renewcommand{\arraystretch}{0.9}
\begin{table*}[t]
    \centering
    \small
    \caption{Evaluation results of \textit{Visual Planning} on VP$^2$~\cite{tian2023control} Benchmark. Top-1 performance across 8 tasks and the average success rate are highlighted accordingly. We provide the mean and standard deviation of the success rate (in \%) on average over 3 runs. The best and second-best performances are represented with \textbf{bold text} and \underline{underlines}, respectively.}
    \vspace{-2pt}
    \renewcommand{\arraystretch}{1.0}
    \setlength{\tabcolsep}{3.5pt}
    \begin{threeparttable}
    \begin{tabularx}{0.99\linewidth}{l *{8}{>{\centering\arraybackslash}m{1.39cm}} | *{2}{>{\centering\arraybackslash}m{0.7cm}}}
        \toprule[1pt]
        Method & Robosuite Push & Flat Block & Open Drawer & Open Slide & Blue Button & Green Button & Red Button & Upright Block & \multicolumn{2}{c}{Avg. Success} \\
        \midrule[1pt]
        Simulator & 93.5$^{\pm2.2}$ & 13.3$^{\pm0.1}$ & 76.7$^{\pm0.0}$ & 71.7$^{\pm1.2}$ & 100.0$^{\pm0.0}$ & 96.7$^{\pm0.0}$ & 90.0$^{\pm0.0}$ & 90.0$^{\pm0.0}$ & 88.4 & * \\
        \midrule[0.5pt]
        MCVD~\cite{voleti2022MCVD} & 77.3$^{\pm2.6}$ & 4.0$^{\pm1.1}$ & 11.7$^{\pm1.2}$ & 18.3$^{\pm1.0}$ & 95.0$^{\pm3.6}$ & 83.3$^{\pm0.4}$ & 73.3$^{\pm2.6}$ & 56.7$^{\pm2.4}$ & 59.4 & 67.2 \\
        FitVid~\cite{babaeizadeh2021fitvid} & 67.7$^{\pm5.3}$ & \textbf{9.2}$^{\pm4.0}$ & 25.3$^{\pm6.9}$ & \textbf{35.3}$^{\pm4.5}$ & 94.0$^{\pm4.6}$ & \underline{84.0}$^{\pm5.3}$ & 58.7$^{\pm5.1}$ & 51.3$^{\pm2.7}$ & 59.5 & 67.3 \\
        MaskViT~\cite{gupta2022maskvit} & \textbf{82.6}$^{\pm2.3}$ & 4.0$^{\pm3.9}$ & 4.0$^{\pm4.5}$ & 8.7$^{\pm5.7}$ & 94.7$^{\pm2.0}$ & 64.0$^{\pm4.3}$ & 24.0$^{\pm7.5}$ & \textbf{62.2}$^{\pm8.6}$ & 48.6 & 55.0 \\
        iVideoGPT\cite{wu2024ivideogpt} & 78.3$^{\pm0.4}$ & 3.3$^{\pm0.7}$ & \underline{37.5}$^{\pm1.5}$ & 16.1$^{\pm2.5}$ & \underline{95.6}$^{\pm2.9}$ & 82.5$^{\pm3.1}$ & \underline{92.2}$^{\pm1.5}$ & 44.7$^{\pm1.7}$ & \underline{63.9} & \underline{72.2} \\
        \rowcolor{gray!20}
        \ourwork (Ours) & \underline{81.4}$^{\pm1.7}$ & \underline{6.1}$^{\pm2.0}$ & \textbf{40.5}$^{\pm1.1}$ & \underline{19.9}$^{\pm3.4}$ & \textbf{96.7}$^{\pm2.5}$ & \textbf{85.6}$^{\pm3.0}$ & \textbf{93.2}$^{\pm1.9}$ & \underline{44.8}$^{\pm1.4}$ & \textbf{66.0} & \textbf{74.7} \\
        \bottomrule[1pt]
    \end{tabularx}

    \begin{tablenotes}
    \footnotesize
    \item[*] Values in this column are normalized by the simulator’s average success rate.
    \end{tablenotes}

    \end{threeparttable}
    \label{tab:vp}
    \vspace{-12pt}
\end{table*}

%% file: sections_cvpr/4_experiment.tex
\vspace{-5pt}
\section{Experiments}
\label{sec:exp}
In this section, we conduct comprehensive experiments to validate \ourwork model on multiple tasks, including \textit{controllable video generation}, \textit{visual planning}, and \textit{policy learning}.
They are expected to answer these questions: \textit{1) What is the quality of the videos generated by \ourwork? 2) To what extent is the generative capability of \ourwork? 3) How can generated videos benefit robot learning tasks?}
Additionally, we provide dataset details, experiment details, and more results in Sec.~\ref{sec:supp_data}, Sec.~\ref{sec:supp_impl}, Sec.~\ref{sec:supp_exp} in Suppl., respectively. 


\begin{figure*}
    \centering
    \includegraphics[width=\textwidth]{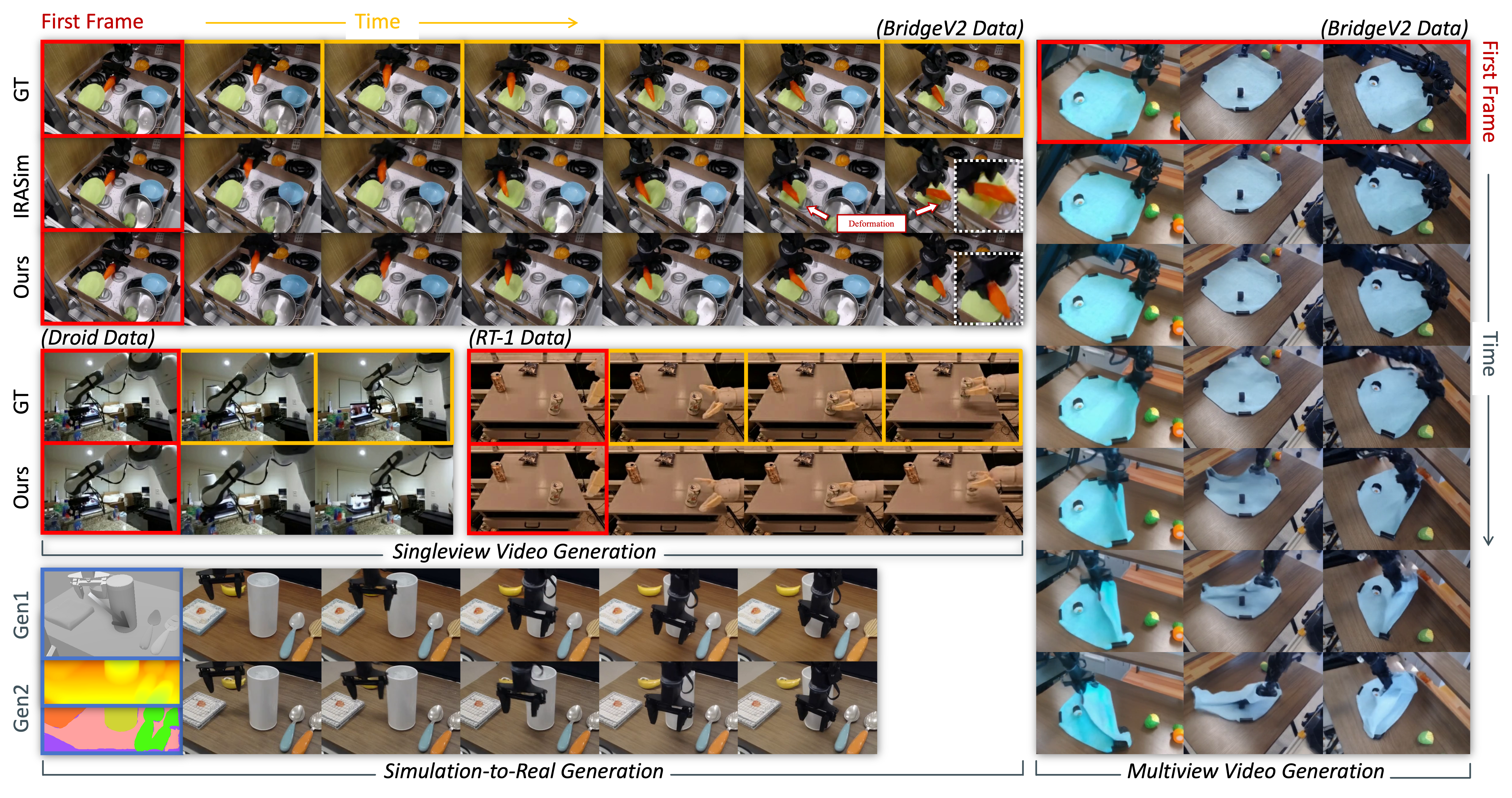}
    \vspace{-15pt}
    \caption{Qualitative results of versatile \textbf{Video Generation} with full conditions. Given one-frame observation, \ourwork predict subsequent 15 frames on validation split of Bridge~\cite{walke2023bridgedata}, DROID~\cite{khazatsky2024droid}, RT-1~\cite{brohan2022rt1} datasets. \textcolor{red}{Red boxes} denotes the first frame input of the video generation; \textcolor{orange}{Orange boxes} denotes the ground-truth of the subsequence frames.}
    \label{fig:video_gen}
    \vspace{-15pt}
\end{figure*}


\subsection{Conditional Video Generation}
\label{sec:exp_video_gen}

\mypar{Setup.}
We evaluate the video generation of \ourwork on three real-world datasets, their embodiments, views of each episode, and volume are as below:
\begin{itemize}
    \item BridgeV2~\cite{walke2023bridgedata}: WidowX, 1$\sim$3 views, $\sim$60K episodes;
    \item DROID~\cite{khazatsky2024droid}: Franka Panda, 2 views, $\sim$76K episodes;
    \item RT-1~\cite{brohan2022rt1}: Google Robot, 1 view, $\sim$120K episodes;
\end{itemize}
Please refer to Sec.~\ref{sec:supp_data} in Suppl. for more dataset details.
For the action-conditioned base model setup, we train \ourwork for $\sim$30K steps from the pretrained backbone.
For occupancy maps-guided finetuning and multiview video generation, we have additional $\sim$20K gradient steps of training.

\mypar{Comparison with Baselines.}
To comprehensively demonstrate the superiority of \ourwork model, we compare \ourwork with original CogVideoX~\cite{yang2024cogvideox} and action-conditioned methods AVID~\cite{rigter2024avid}, HMA~\cite{wang2025hma}, IRASim~\cite{zhu2024irasim} and more baselines augmented with our occupancy priors (\textit{e.g.}, IRASim).
We report the quantitative results of \textit{controllable video generation} in Table~\ref{tab:video_gen}, where \ourwork outperforms all baselines across most of the metrics.
Moreover, as highlighted (white arrows) in the BridgeV2 example of the singleview generation in Fig.~\ref{fig:video_gen}, the baseline fails to faithfully infer the dynamics of objects manipulated by the robotic gripper.
More details about the baselines and comparison results are in Sec.~\ref{sec:supp_exp_gen} in Suppl.


\mypar{Multiview Robot Video Generation.}
We show an example of multiview robot video generation performed by \ourwork in Fig.~\ref{fig:video_gen}.
The example shows the robot arm performing a cloth-folding task across \textit{three} views, where the outputs maintain exceptional cross-view consistency.
This high-fidelity multi-view generation enables efficient downstream applications, including photorealistic scene reconstruction and robotics imitation learning.
Note that there exists a lighting discrepancy issue in the original data.

\mypar{Sim-to-Real Transfer.}
Fig.~\ref{fig:video_gen} illustrates examples of sim-to-real generation through \ourwork-S2R, as described in Sec.~\ref{sec:sim2real}.
Details of simulation environment setup and dynamics data generation are provided in Sec.~\ref{sec:supp_s2r} of the Supplementary.
Leveraging an additional image generator (ControlNet~\cite{zhang2023adding}), we first produce diverse initial frames and then extend them to high-quality, realistic manipulation videos.
In this case, using simulator-derived occupancy maps consistent with training preserves the consistent performance.
Moreover, thanks to the robustness of occupancy representations, even condition maps with various granularity (e.g., parametric maps from the simulator) yield only minor performance degradation (see results in Sec.~\ref{sec:exp_abla}).

\vspace{-3pt}
\subsection{Visual Planning}

\mypar{Setup.}
We further evaluate the controllability of \ourwork on VP$^2$~\cite{tian2023vp2}, a visual planning by action controls benchmark.
Following~\cite{tian2023control,wu2024ivideogpt,wang2025sampo}, we train \ourwork on 5K trajectories for Robosuite~\cite{robosuite2020} and 35K for RoboDesk~\cite{kannan2021robodesk}.

\renewcommand{\arraystretch}{0.95}
\begin{table}[t]
\centering
\caption{Evaluation results on SimplerEnv-WidowX~\cite{li2024evaluating} across four manipulation tasks. ``+Finetune'' indicates the additional finetuning on domain-specific dataset; and ``+ORV'' indicates that we augment the finetuning dataset with \ourwork-synthesized data.}
\vspace{-5pt}
\begin{threeparttable}
\resizebox{0.99\linewidth}{!}{
    \begin{tabular}{r | c c c c | c}
    \toprule[1.5pt]
    Method & \makecell{Spoon\\on Towel} & \makecell{Carrot\\on Plate} & \makecell{Stack\\Cube} & \makecell{Eggplant\\in Basket} & \makecell{Avg.\\Success} \\
    \midrule[1pt]
    RoboVLM$^{*\dagger}$~\cite{liu2025towards} & 18.6\% & 22.9\% & 8.1\% & 0.0\% & 12.4\% \\
    +Finetune$^{*}$ & 27.6\% & 26.7\% & 12.1\% & 52.8\% & 29.8\% \\
    \rowcolor{gray!10}
    +\ourwork & 32.2\% & 29.6\% & 15.7\% & 57.9\% & \textbf{33.9\%} \\
    \rowcolor{gray!10}
    \textit{$\Delta$Improvement} & \textcolor{Black}{\textit{+4.6\%}} & \textcolor{Black}{\textit{+2.9\%}} & \textcolor{Black}{\textit{+3.6\%}} & \textcolor{Black}{\textit{+5.1\%}} & \textcolor{Black}{\textbf{\textit{+4.1\%}}} \\
    \midrule[0.5pt]
    SpatialVLA$^{*\dagger}$~\cite{qu2025spatialvla} & 12.5\% & 20.8\% & 20.8\% & 58.3\% & 28.1\% \\
    +Finetune$^{*}$ & 12.8\% & 26.1\% & 26.5\% & 79.3\% & 36.2\% \\
    \rowcolor{gray!10}
    +\ourwork & 14.7\% & 28.4\% & 27.8\% & 83.0\% & \textbf{38.5\%} \\
    \rowcolor{gray!10}
    \textit{$\Delta$Improvement} & \textcolor{Black}{\textit{+1.9\%}} & \textcolor{Black}{\textit{+2.3\%}} & \textcolor{Black}{\textit{+1.3\%}} & \textcolor{Black}{\textit{+3.7\%}} & \textcolor{Black}{\textbf{\textit{+2.3\%}}} \\
    \bottomrule[1.5pt]
    \end{tabular}
}

\begin{tablenotes}
\footnotesize
\item[*] The results are reproduced locally for fully fair comparisons.
\item[$\dagger$] Zero-shot performance (Pretraining).
\end{tablenotes}
\end{threeparttable}

\captionsetup{font=footnotesize}
\vspace{-20pt}
\label{tab:policy_bridge}
\end{table}

\mypar{Results.}
Tab.~\ref{tab:vp} presents the success rates of \ourwork compared to the baselines over 9 tasks.
\ourwork outperforms all baselines in four RoboDesk tasks and achieves second-best results in the other four tasks, indicating its capability to predict \textit{high-fidelity} future observations, which is fully \textit{controllable}.\looseness=-1

\subsection{Policy Learning}
\label{sec:exp_policy}

\begin{figure*}
    \centering
    \includegraphics[width=\textwidth]{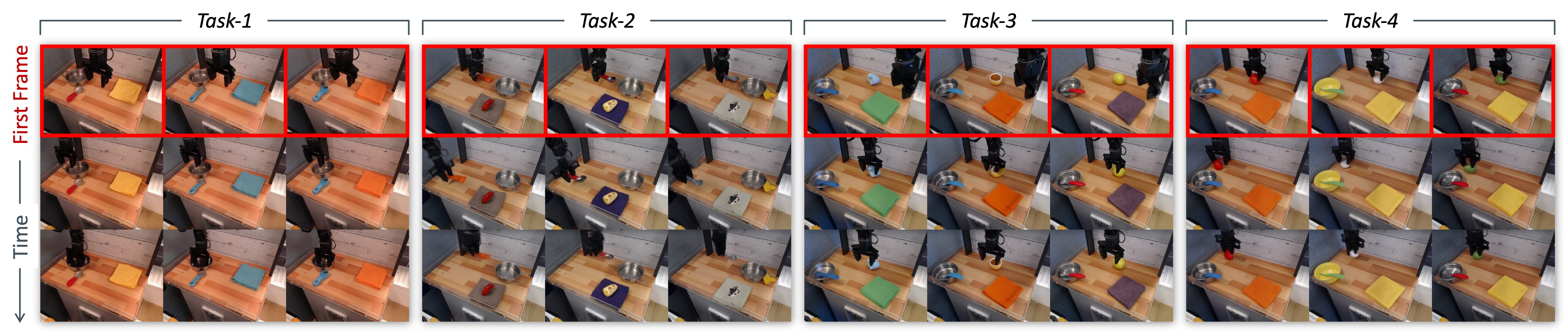}
    \vspace{-15pt}
    \caption{Illustrations of \textbf{Appearance Randomization} powered by \ourwork, generating diverse manipulation videos of four tasks. For each manipulation task in Bridge Data~\cite{walke2023bridgedata}, we present three examples with distinct visual appearances, demonstrating that \ourwork generalizes well to varied context inputs and thereby alleviates the challenge of data collection of robot learning.}
    \label{fig:diverse}
    \vspace{-12pt}
\end{figure*}

\mypar{Setup.}
To improve policy learning, we employ it as a powerful data engine to augment existing data.
Similar to \ourwork-S2R, we leverage another image generator (ControlNet) to generate diverse initial frames and then extend them to videos, with some examples with appearance randomizations are shown in Fig.~\ref{fig:diverse}.
For our evaluations, we use post-finetuning after cross-embodiment pre-training as the setup, and leverage \ourwork to generate additional $\sim30K$ samples (refer to Sec.~\ref{sec:supp_model_policy} in Suppl. for more details).
We evaluate the recent open-sourced policy models RoboVLM~\cite{liu2025towards} and SpatialVLA~\cite{he2015spatial} on SimplerEnv-WidowX~\cite{li2024evaluating} with BridgeData V2~\cite{walke2023bridgedata} as the test suite.
Each policy model is finetuned both on the original data and the augmented data, following the official instructions.
For more discussions about the data augmentation, please refer to Sec.~\ref{sec:supp_data_aug} in Suppl.

\renewcommand{\arraystretch}{0.95}
\begin{table}[t]
\centering
\caption{Ablation results of \textit{Video Generation} and \textit{Visual Planning} on approaches of priors injection.}
\vspace{-5pt}
\resizebox{0.99\linewidth}{!}{
    \begin{tabular}{l | c c c c | c}
    \toprule[1.5pt]
    Variants & PSNR$\uparrow$ & SSIM$\uparrow$ & FID$\downarrow$ & FVD$\downarrow$ & Success$\uparrow$ \\
    \midrule[1pt]
    CogVideoX & 19.432 & 0.752 & 7.509 & 83.561 & - \\
    \midrule[0.5pt]
    \multicolumn{6}{l}{\textit{Action Conditions}} \\
    w/ Text Expert & 20.424 & 0.772 & 4.104 & 23.586 & 52.9 \\
    No Chunks & 24.813 & 0.850 & 3.793 & 19.944 & 70.6 \\
    \rowcolor{gray!10}
    Ours (base) & 25.631 & 0.873 & 3.821 & 17.682 & 74.7\\
    \midrule[0.5pt]
    \multicolumn{6}{l}{\textit{Occupancy Map Conditions}} \\
    ControlNet & 26.974 & 0.865 & 3.613 & 20.069 & - \\
    \rowcolor{gray!10}
    Ours (full) & 28.258 & 0.899 & 3.418 & 16.525 & -\\
    \bottomrule[1.5pt]
    \end{tabular}
}
\captionsetup{font=footnotesize}
\vspace{-15pt}
\label{tab:abla_cond_inject}
\end{table}

\mypar{Results.}
Tab.~\ref{tab:policy_bridge} shows that the augmented data from \ourwork improves policy learning performance.
We keep the original and augmented finetuning data the same in size, with the synthetic data accounting for $\sim$25\% of the augmented data as a practical choice.
With the augmentation (the row of ``+ORV'' in Tab.~\ref{tab:policy_bridge}), we achieve gains of $\sim$13.7\% (29.8\% $\rightarrow 33.9\%$) for RoboVLM~\cite{liu2025towards} and $\sim$6.5\% (36.2\% $\rightarrow 38.5\%$) for SpatialVLA~\cite{qu2025spatialvla}, which significantly demonstrates the effectiveness of \ourwork-augmented data for policy learning.

\renewcommand{\arraystretch}{0.95}
\begin{table}[t]
\centering
\caption{Ablation results of \textit{Conditional Video Generation} on occupancy conditioning resources and training strategies.}
\vspace{-5pt}
\resizebox{0.99\linewidth}{!}{
    \begin{tabular}{c | c c c c c}
    \toprule[1.5pt]
    Variants & Source & PSNR$\uparrow$ & SSIM$\uparrow$ & FID$\downarrow$ & FVD$\downarrow$ \\
    \midrule[1pt]
    \rowcolor{blue!5}
    \multicolumn{6}{c}{\textit{Conditioning Resources}} \\
    w/o cond. (base) & - & 25.631 & 0.873 & 3.821 & 17.682 \\
    \midrule[0.5pt]
    \multirow{2}{*}{w/ depth} & Fine & 30.288 & 0.919 & 3.061 & 14.321 \\
      & Coarse & 28.031 & 0.896 & 4.522 & 18.548 \\
    \midrule[0.5pt]
    \multirow{2}{*}{w/ sem.} & Fine & 28.896 & 0.901 & 3.259 & 16.171 \\
      & Coarse & 27.911 & 0.896 & 3.467 & 17.053 \\
    \midrule[0.5pt]
    \multirow{2}{*}{Full cond.} & \cellcolor{gray!10}Fine & \cellcolor{gray!10}30.431 & \cellcolor{gray!10}0.920 & \cellcolor{gray!10}2.998 & \cellcolor{gray!10}14.301 \\
      & \cellcolor{gray!10}Coarse & \cellcolor{gray!10}28.258 & \cellcolor{gray!10}0.899 & \cellcolor{gray!10}3.418 & \cellcolor{gray!10}16.525 \\
    \midrule[1pt]
    \rowcolor{red!5}
    \multicolumn{6}{c}{\textit{Training Strategies (w/o Occupancy Conditionings)}} \\
    From scratch & - & 23.518 & 0.811 & 19.357 & 84.831 \\
    From CogVideoX2B & \cellcolor{gray!10}- & \cellcolor{gray!10}25.631 & \cellcolor{gray!10}0.873 & \cellcolor{gray!10}3.821 & \cellcolor{gray!10}17.682 \\
    \bottomrule[1.5pt]
    \end{tabular}
}
\captionsetup{font=footnotesize}
\vspace{-15pt}
\label{tab:abla_cond}
\end{table}

\vspace{-2pt}
\subsection{Ablation Study and Analysis}
\label{sec:exp_abla}
\vspace{-3pt}

In this section, we present comprehensive ablations of our proposed \ourwork framework and other related discussions.

\mypar{Effect of Conditioning Approaches.}
We first ablate the action conditioning designs in Fig.~\ref{fig:mod} with the results shown in Tab.~\ref{tab:abla_cond_inject}.
Different configurations of the Action Expert AdaLN (\textit{e.g.}, take the combination of original Vision Expert and Text Expert) result in significantly inferior performance.
In addition, using action conditioning without temporal chunking (\textit{e.g.}, directly encoding the entire action sequence) also weakens the performance (PSNR drops by 3.2\% and success rate drops by 5.5\%).
For occupancy-derived visual conditionings, we validate the effectiveness of injecting the conditioning into the initial noise.
We confirm that injecting occupancy-derived coarse controls into deep layers causes noticeable performance degradation (PSNR drops by 4.5\%).

\mypar{Effect of Control Signals.}
Tab.~\ref{tab:abla_cond} reveals the impact of different conditioning resources (Coarse: occupancy-rendered condition maps; Fine: pixel-level condition maps) and conditioning types (depth and semantic) used for training and evaluation.
The results demonstrate that introducing visual priors leads to significant improvements, with gains of 18.72\% (25.621$\rightarrow$30.431) and 10.24\% (25.621$\rightarrow$28.258).
Moreover, coarse condition maps achieve performance comparable to their fine counterparts.
In addition, Tab.~\ref{tab:abla_mv_cond} further shows the improvements in three-view (BridgeData V2~\cite{walke2023bridgedata}) robot video generation when visual priors are introduced, where the view0 serves as the ``anchor view'' for constructing multiview visual priors (see Sec.~\ref{sec:supp_mv} in Suppl.).

\mypar{Effect of Pretraining.}
We further test the benefits of the pretraining process. As shown in Tab.~\ref{tab:abla_cond}, models trained from the CogVideoX have superior performance compared to those from scratch, particularly on FID and FVD metrics.

\renewcommand{\arraystretch}{0.95}
\begin{table}[t]
\centering
\caption{Ablation results of \textit{Multiview Video Generation} on occupancy conditionings on BridgeData V2~\cite{walke2023bridgedata} with 3 views. Numbers are reported as ``with / without'' visual priors.}
\vspace{-5pt}
\resizebox{0.99\linewidth}{!}{
    \begin{tabular}{c | c c c c}
    \toprule[1.5pt]
    Views & PSNR$\uparrow$ & SSIM$\uparrow$ & FID$\downarrow$ & FVD$\downarrow$ \\
    \midrule[1pt]
    View0 (anchor) & 25.77 / 28.25 & 0.87 / 0.89 & 3.20 / 3.11 & 14.05 / 12.54 \\
    View1 & 23.04 / 25.87 & 0.79 / 0.85 & 3.31 / 3.18 & 16.36 / 13.67 \\
    View2 & 22.90 / 25.79 & 0.78 / 0.85 & 3.32 / 3.19 & 15.97 / 13.62 \\
    \bottomrule[1.5pt]
    \end{tabular}
}
\captionsetup{font=footnotesize}
\vspace{-5pt}
\label{tab:abla_mv_cond}
\end{table}

\renewcommand{\arraystretch}{0.95}
\begin{table}[t]
\centering
\caption{Ablation results of zero-shot \textit{Conditional Video Generation} on different occupancy conditioning resources.}
\vspace{-5pt}
\resizebox{0.99\linewidth}{!}{
    \begin{tabular}{c c | c c c c}
    \toprule[1.5pt]
    Train & Val & PSNR$\uparrow$ & SSIM$\uparrow$ & FID$\downarrow$ & FVD$\downarrow$ \\
    \midrule[1pt]
    \rowcolor{gray!10}
    Coarse & Coarse & 28.031 & 0.896 & 4.522 & 18.548 \\
    \rowcolor{gray!10}
    Coarse & Fine & 26.608 {\scriptsize\textcolor{BrickRed}{(-1.423)}} & 0.872 {\scriptsize\textcolor{BrickRed}{(-0.024)}} & 4.932 {\scriptsize\textcolor{BrickRed}{(+0.410)}} & 24.134 {\scriptsize\textcolor{BrickRed}{(+5.586)}} \\
    \midrule[0.5pt]
    Fine & Fine & 30.288 & 0.919 & 3.061 & 14.321 \\
    Fine& Coarse & 19.048 {\scriptsize\textcolor{BrickRed}{(-11.240)}} & 0.754 {\scriptsize\textcolor{BrickRed}{(-0.165)}} & 22.893 {\scriptsize\textcolor{BrickRed}{(+19.832)}} & 132.685 {\scriptsize\textcolor{BrickRed}{(+109.792)}} \\
    \bottomrule[1.5pt]
    \end{tabular}
}
\captionsetup{font=footnotesize}
\vspace{-13pt}
\label{tab:abla_zero_shot}
\end{table}

\mypar{Robustness of Occupancy Representations.}
To validate the robustness of occupancy representations used in \ourwork model, as described in Sec.~\ref{sec:formulation} and Sec.~\ref{sec:exp_video_gen}.
We examine the zero-shot performance of models through training and evaluate them under different condition settings, as illustrated in Tab.~\ref{tab:abla_zero_shot} (refer to Fig.~\ref{fig:abla_fine_coarse} in Suppl. for more qualitative details).
The results reveal that models trained on occupancy-derived coarse visual conditions generalize better across conditions of varying granularity.
In contrast, \ourwork models trained on pixel-aligned conditions suffer a dramatic performance drop on coarse inputs.
This imposes a major constraint on deploying the model in more diverse scenarios (\textit{e.g.}, from simulation to real-world), necessitating condition maps that accurately align with ground truths.
Therefore, previous works~\cite{alhaija2025cosmos,liu2025robotransfer} are sensitive to inaccuracies in the conditioning, whereas \ourwork is not.

\mypar{More Discussions.}
We have additional broader discussions for a better understanding of our work in Sec.~\ref{sec:supp_discuss} in Suppl.

%% file: sections_cvpr/5_conclusion.tex
\vspace{-5pt}
\section{Conclusion}
\vspace{-5pt}

We propose \ourwork, an occupancy-centric framework for robot video generation that couples action priors with occupancy-derived visual priors.
With such an occupancy-centric design, \ourwork achieves high-quality robot video generation and consistent multiview synthesis.
The robustness of occupancy representations further enables \ourwork to achieve superior visual transfer between simulated and real-world scenarios.
Experiments on controllable video generation, visual planning, and policy learning demonstrate the effectiveness and versatility of \ourwork for advancing robotics research.

%% file: sections_cvpr/X_suppl.tex
\clearpage
\setcounter{page}{1}
\maketitlesupplementary




\vspace{10mm}

\noindent Our supplementary contains following contents:

\begin{enumerate}[label=(\Alph*), align=left]
    \item \textbf{Demo Video.} We provide more illustrative videos to demonstrate the motivation and demos in Sec.~\ref{sec:supp_demo}.
    \item \textbf{Dataset Details.} In addition to the key components introduced, we describe other modules of \ourwork in Sec.~\ref{sec:supp_data}.
    \item \textbf{ORV-MV Details.} We have more details of how we build \ourwork-MV model (\textit{e.g.}, training data) in Sec.~\ref{sec:supp_mv}.
    \item \textbf{ORV-S2R Details.} We explain how we build simulation-to-real framework \ourwork-S2R in Sec.~\ref{sec:supp_s2r}.
    \item \textbf{Implementation Details.} We provide other all details regarding the implementations, training and evaluation, for the purpose of reproducing, in Sec.~\ref{sec:supp_impl}.
    \item \textbf{Additional Results.} We have more experiments and analysis in Sec.~\ref{sec:supp_exp}.
    \item \textbf{Discussions.} We have broader range of discussions including the concurrent related works, limitations and the potential improvements of \ourwork in Sec.~\ref{sec:supp_discuss}.
    \item \textbf{License.} We list licenses of all assets used in \ourwork. 
\end{enumerate}

\begin{figure*}
    \centering
    \includegraphics[width=\textwidth]{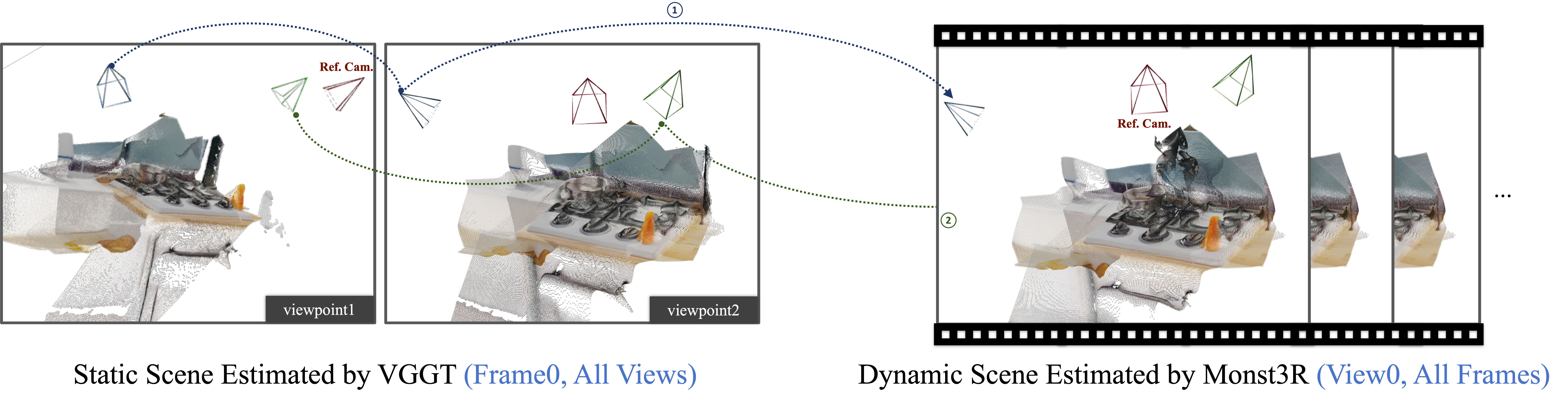}
    \vspace{-20pt}
    \caption{Illustration of \ourwork aligning multiview cameras from VGGT~\cite{wang2025vggt} under the frame of MonST3R~\cite{zhang2024monst3r} to get the multiview conditioning sequences.}
    \label{fig:mv_cam}
    \vspace{-15pt}
\end{figure*}

\section{Supplementary Video}
\label{sec:supp_demo}

We provide additional videos for better demonstration of \boldourwork.
These videos showcase high-quality conditional robot video generation that closely resemble the ground truth.
We also include videos of multiview video generations.
Note that all videos are muted.
Please refer to the attached anonymous page file (\texttt{index.html}) for the details.

\section{Datasets Details}
\label{sec:supp_data}

\mypar{BridgeData V2~\cite{walke2023bridgedata}} is a large-scale, diverse collection of robot manipulation data in real-world robotic platforms. It includes 60096 trajectories, spanning 24 various environments and a wide range of tasks (\textit{e.g.}, pushing, placing, opening, and insertion). In our experiments, we use the version of $480\times640$ (Raw data) for the singleview training and evaluations (keep aligned with the baselines), while use the version of $256\times256$ (RLDS data) for the multiview training and evaluation. BridgeV2 also offers the 7-DoF action and language labels.

\mypar{DROID~\cite{khazatsky2024droid}} has nearly 76K teleoperated trajectories ($\sim$350 hours) spanning 86 tasks in 564 scenes. It includes multiview (2 side views and 1 wrist view) RGB, depth 7DoF action labels, and language instructions. In our experiments, we use the version of $180\times320$ (RLDS data) for all the training and evaluations.

\mypar{RT-1~\cite{brohan2022rt1}} is a large-scale real-world robot manipulation dataset of over 130K trajectories collected in office-like environments. Each episode is paired with RGB observation, 7DoF action, and language labels, across diverse tasks such as picking, placing, and opening. In our experiments, we use the version of $256\times320$ for all the training and evaluations.


\section{\ourwork-MV Details (Section~\ref{sec:mv})}
\label{sec:supp_mv}

In Fig.~\ref{fig:mv_pipeline}, we use the multiview 2D conditioning maps to guide the multiview video generations, just as we do in singleview video generations~\ref{sec:vgm}.
However, giving that no well-prepared or publicly available camera parameters data are released in our adapted dataset, we provide more details about how we get such data in our model training.

As described in Sec.~\ref{sec:data}, we extract 4D points from a singleview input (referred to as the ``anchor view'' or ``reference view'') using MonST3R~\cite{zhang2024monst3r}.
To get multiview conditions, we estimate camera poses across all views in the dataset using VGGT~\cite{wang2025vggt}.
Note, however, that the two estimation approaches produce different coordinate spaces for the 4D points and camera poses.

\begin{figure*}
    \centering
    \includegraphics[width=\textwidth]{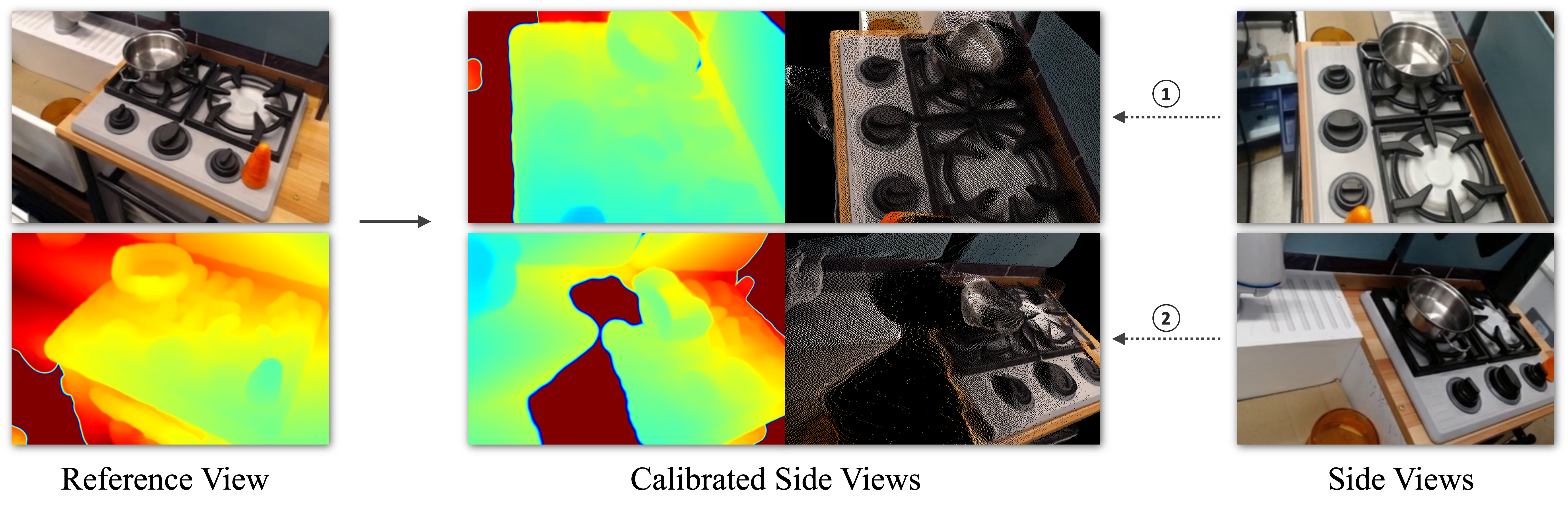}
    \vspace{-20pt}
    \caption{Example of transferring multiview poses from VGGT~\cite{wang2025vggt} to MonST3R~\cite{zhang2024monst3r}. The comparison of calibrated side views and the side views demonstrates the efficiency.}
    \label{fig:demo_cam}
    \vspace{-10pt}
\end{figure*}

We then have a simple yet efficient approach to combine the advances of MonST3R~\cite{zhang2024monst3r} and VGGT~\cite{wang2025vggt}.
As illustrated in Fig.~\ref{fig:mv_cam}, these two reconstruction methods share a common rule: they both take the first frame (of MonST3R) or the first view (of VGGT) as their reference coordinate space. Hence, we perform efficient pixel-wise matching on the first frame (view) to extract the global \textit{scale} ($\alpha$) and \textit{shift} ($\beta$) vectors, which enables the reciprocal transformation between the two coordinate spaces.
In such a way, we can add all other calibrated cameras in the frame of MonST3R.
Specifically, we apply the Linear-Least-Squares Fitting~\cite{bjorck2024numerical} on the depth maps to estimate these values~\cite{yu2022monosdf}, as Eq.~\ref{eq:lls}:
\begin{equation}
\label{eq:lls}
\mathrm{Solve:} \min_{\alpha, \beta} \sum_{i \in \mathcal{V}} \left( \alpha D^{\prime}_i + \beta - D_i \right)^2,
\end{equation}
where $\mathcal{V}$ means the image space, $D$ and $D^{\prime}$ denote the reference depth map from MonST3R and VGGT, respectively.
More efficiently, we omit the shift and use the \textit{scale} solely in our practice---again because the exactly identical reference coordinate space is shared, and given that the predicted 3D points from both approaches do not exhibit significant offset errors.
Fig.~\ref{fig:demo_cam} shows an example of the camera poses alignment by simply estimating the \textit{scale} vector.
Given the reconstructed 4D points (occupancy) from the reference view, we can render the conditioning sequences from all views (reference view + calibrated side views).

\section{\ourwork-S2R Details (Section~\ref{sec:sim2real})}
\label{sec:supp_s2r}

For the results shown in Sec.~\ref{sec:exp_video_gen}, our simulated tabletop manipulation environments are constructed within the ManiSkill~\cite{mu2021maniskill} framework.
We aim to utilize the efficient simulator to generate the simulated dynamics data with corresponding geometries (\textit{e.g.}, mesh and occupancy), based on which \ourwork will further generate the realistic manipulation data of diverse scenarios.

\begin{figure}
\centering
\includegraphics[width=\linewidth]{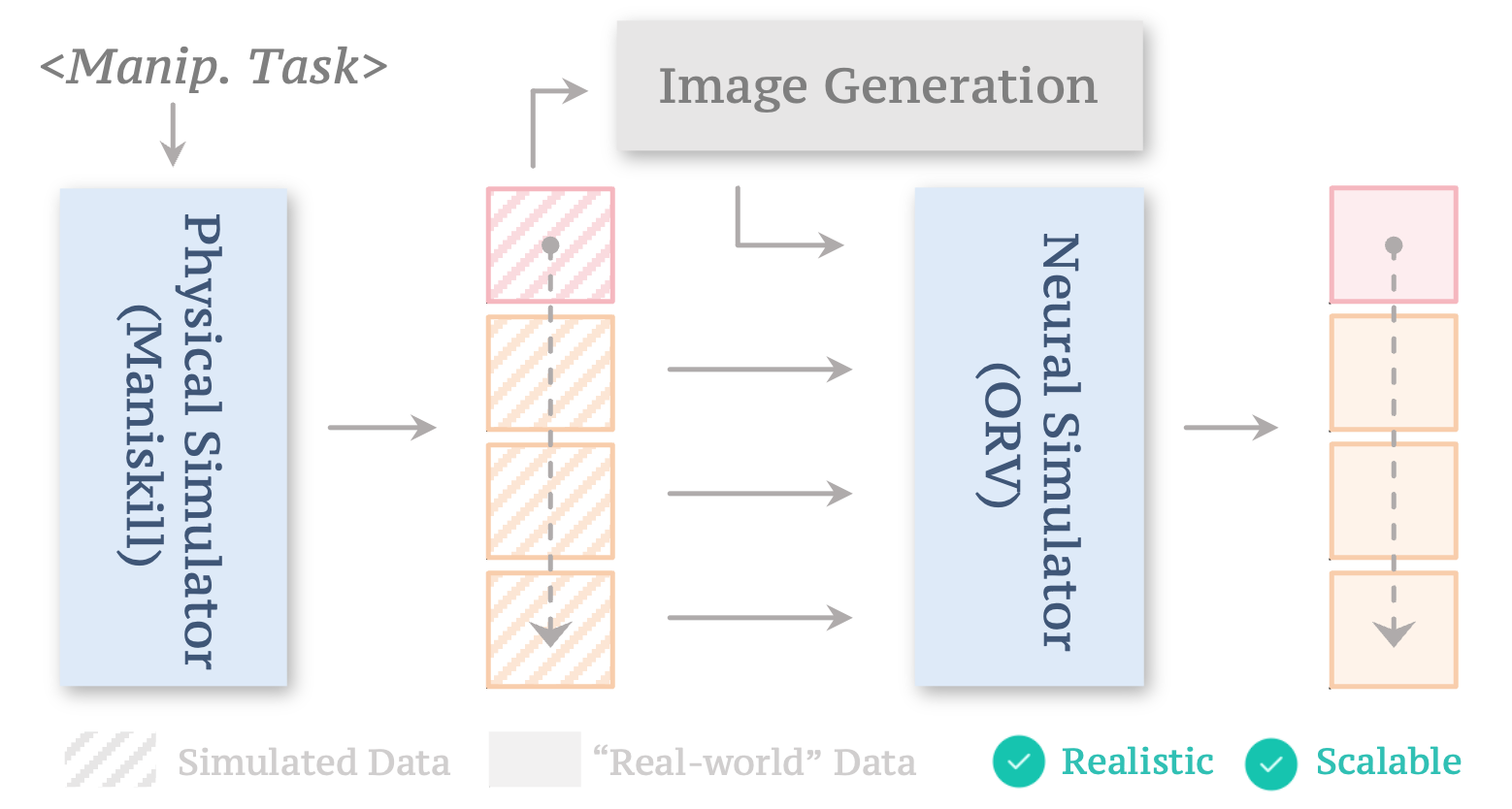}
\captionsetup{font=footnotesize}
\caption{Illustration of Simulation-to-Real Generation of \ourwork.}
\vspace{-15pt}
\label{fig:sim2real_pipe}
\end{figure}

Previous work SIMPLER~\cite{li2024evaluating} shares similar thoughts, as SIMPLER claims that simulation-based evaluation can be a scalable, reproducible and reliable proxy for real-world evaluation.
However, the difference is that SIMPLER focuses on duplicating the real-world policy evaluation in a simulation environment and mitigating the gaps of dynamics transfer.
While our primary objective is the sim-to-real visual transfer.
Specifically, it involves constructing tabletop scenes within the ManiSkill, followed by structured object placement and policy-driven interaction.
We first collect diverse 3D assets from public datasets or even enrich them with reconstructed objects from 2D images~\cite{ye2025hi3dgen,tochilkin2024triposr}.
Objects are placed on predefined tabletop regions using a grid-based sampling strategy to ensure diverse yet physically plausible layouts.
To enable meaningful interactions, we train reinforcement-learning policies inspired by UniGraspTransformer~\cite{wang2024unigrasptransformer} for object-specific grasping.
Executing these policies produces rich trajectories across varied scenes, from which spatial occupancy data are systematically generated to condition our \ourwork model.

To complete the simulation-to-real visual transfer described in Sec.~\ref{sec:sim2real}, we simply leverage the ControlNet~\cite{zhang2023adding} (depth-to-image) trained from x-flux\footnote{\href{https://github.com/XLabs-AI/x-flux}{https://github.com/XLabs-AI/x-flux}} release to synthesize initial frames.
By conditioning on depth and semantic maps, the appearance of these frames can be flexibly controlled through text instructions or just multiple runs with different seeds, as illustrated in Fig.~\ref{fig:sys} and Fig.~\ref{fig:video_gen}.
Combined with the generalization ability of \ourwork, diverse visual renditions of the same manipulation task can be generated, effectively supporting data augmentation for robot policy learning.

\section{Implementation Details}
\label{sec:supp_impl}

We provide more details regarding the implementation of our dataset curation, methods and experiments, including all the empirical hyperparameters and settings.

\subsection{Occupancy Dataset Curation (Section~\ref{sec:data})}
\label{sec:supp_occ_data}

\subsubsection{Data Construction}

\mypar{Semantics Labels.} In the process of dataset-level semantics labelset construction, we employ the VLM (Qwen-VL-Chat\footnote{\href{https://huggingface.co/Qwen/Qwen-VL-Chat}{https://huggingface.co/Qwen/Qwen-VL-Chat}}~\cite{bai2023qwen}) to exhaustively caption all the scenarios in the dataset (as~\cite{wang2025semantic,shuai2025pugs}). Specifically, we use the text instruction as below. 
To construct a compact yet representative label set that covers most labels in the dataset, we embed all $\sim$150K extracted labels using \texttt{all-MiniLM-L6-v2}\footnote{\href{https://huggingface.co/sentence-transformers/all-MiniLM-L6-v2}{https://huggingface.co/sentence-transformers/all-MiniLM-L6-v2}}~\cite{wang2020minilmv2}
 and apply K-Means clustering to the resulting embeddings with the number of clusters set to 51.\looseness=-1

\begin{tcolorbox}[
colback=gray!10, 
colframe=gray!40, 
boxrule=0.4pt, 
arc=0.2mm, 
left=4pt, right=4pt, top=3pt, bottom=3pt, width=\linewidth, ] \small\ttfamily List the main object classes in the image, with only one word for each class:
\end{tcolorbox}

\mypar{Occupancy.} In the process of points-to-occupancy transformation, we adjust the voxel size to get the trade-off between the computation cost and the granularity of the geometry surface. Specifically, we use a voxel size of $0.001^3$ units. The overall spatial extent is set to $0.4 \times 0.4 \times 0.4$ units for the BridgeV2 dataset, and $0.4 \times 0.4 \times 0.6$ units for the DROID and RT-1 datasets.

\subsubsection{Rendering with Adaptive Scaling}
\label{sec:supp_data_render}

As described in Sec.~\ref{sec:data}, we apply a adaptive scaling rule $\sigma=k_2\cdot \hat{z}^{k_1}$ on the size of Gaussian splattings, with an exponential term $k_1$ and a base scale term $k_2$.

\mypar{Exponential Term $k_1$.} When the Gaussian center is at depth $z$ under camera coordinate space, its rendered standard deviation on the image plain is approximately $\sigma_{\text{img}}\approx \tfrac{f}{z}\sigma_{\text{cam}}$, where f denotes the focal length in pixels and $\sigma_{\text{cam}}$ is the Gaussian scale in 3D space.
Consequently, the projected pixel area of a Gaussian follows a simple quadratic inverse relation with depth: $a_{\text{img}}\propto (\tfrac{1}{z})^2$.
In this case, using a fixed Gaussian scale $\sigma$ during rendering results in distorted appearances: Gaussians closer to the camera occupy larger image regions, whereas distant ones shrink rapidly, with their rendered area decreases \textit{exponentially} with depth.
This observation naturally motivates the exponential term $z^{k_1}$ in our scaling schedule.

\mypar{Base Scale Term $k_2$.}
Since $(\tfrac{1}{z})^2$ exhibits opposite variation rates when $z < 1$ and $z > 1$, the exponential term $z^{k_1}$ exerts an increasingly strong influence on the rendered area $a_{\text{img}}$ as $z\!\rightarrow\!0$ or $z\!\rightarrow\!\infty$.
This leads to a two-pole issue—no single optimal choice of $k_1$ can simultaneously balance both extremes.
To mitigate this, we normalize the depth range to $\hat{z}\!\in\![1, 2)$ in a canonical space: $\hat{z}=(z-\text{min}(z))/(\text{max}(z)-\text{min}(z))+1$, leaving only one pole ($z\!\rightarrow\!\infty$) corresponding to Gaussians far from the image plane.
An additional base term $k_2$ is then introduced to control the scale of Gaussians near the image plane ($z\!\rightarrow\!0,\hat{z}\!\rightarrow\!1$), ultimately yielding the adaptive scaling rule: $\sigma=k_2\cdot \hat{z}^{k_1}$.

In our experiments, we adapt the implementation from \texttt{diff-gaussian-rasterization}\footnote{\href{https://github.com/graphdeco-inria/diff-gaussian-rasterization}{https://github.com/graphdeco-inria/diff-gaussian-rasterization}}. We set $k_1=3.7$, $k_2=0.00023$ for the BridgeData V2~\cite{walke2023bridgedata} dataset, and $k_1=3.2$, $k_2=0.00047$ for DROID~\cite{khazatsky2024droid} and RT-1~\cite{brohan2022rt1} datasets.

\begin{figure}
\centering
\includegraphics[width=\linewidth]{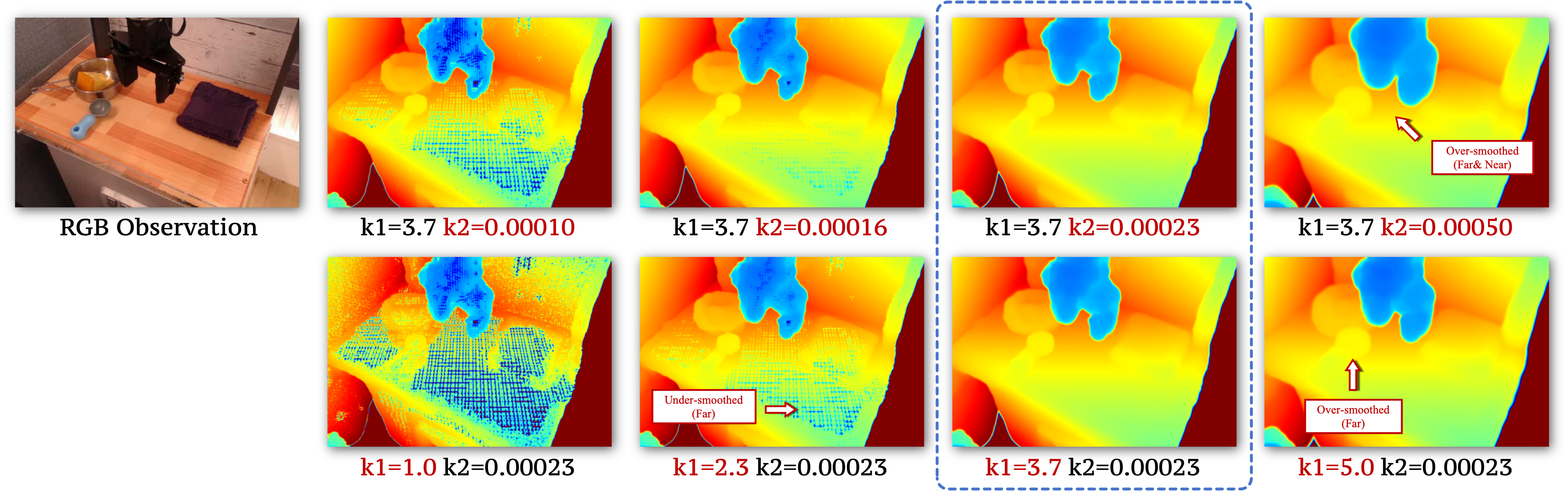}
\captionsetup{font=footnotesize}
\vspace{-15pt}
\caption{Rendering examples of different choices of $k_1,k_2$ on BridgeData V2~\cite{walke2023bridgedata}. The blue box marks the empirically chosen value in our implementation. Better to zoom in.}
\vspace{-10pt}
\label{fig:gs_scale}
\end{figure}

Fig.~\ref{fig:gs_scale} showcases examples of different combinations of $k_1$ and $k_2$ during rendering, where we can observe that $k_1$ has a stronger influence on Gaussians far from the image plane, while those near the plane are mainly adjusted by $k_2$. We empirically determine the optimal values through exhaustive enumeration as highlighted by blue box.

\begin{table}[!ht]
\renewcommand{\arraystretch}{0.95}
\centering
\caption{Main hyperparameters of model architecture, where $^*$ denotes those that are specialized in our model, while others keep the same as the CogVideoX-2B.}
\vspace{-10pt}
\vspace{5pt}
\label{tab:supp_model}
\setlength{\tabcolsep}{3.5pt}
\begin{threeparttable}
\begin{tabular}{l >{\centering\arraybackslash}p{2cm}}
\toprule[1.5pt]
    Hyperparameter & Value \\
    \midrule[1pt]
    \multicolumn{1}{l}{\textit{Model}} & \\
    input channels & 32\rlap{\raisebox{2pt}{\tnote{*}}} \\
    attention head dimension & 64 \\
    number of attention heads & 30 \\
    number of transformer blocks & 30 \\
    output channels & 16 \\
    patch size & 2 \\
    text embedding dimension & 4096 \\
    diffusion timestep embedding dimension & 512 \\
    action embedding dimension & 512\rlap{\raisebox{2pt}{\tnote{*}}} \\
    conditioning dimension & 1920\rlap{\raisebox{2pt}{\tnote{*}}} \\
    positional encoding & sin,cos \\
    \midrule[0.5pt]
    \multicolumn{1}{l}{\textit{VAE}} & \\
    spatial compression ratio & 8 \\
    temporal compression ratio & 4 \\
\bottomrule[1.5pt]
\end{tabular}
\vspace{-5pt}
\end{threeparttable}
\end{table}

\begin{figure*}[t]
\centering
\begin{minipage}{0.92\textwidth}
\begin{lstlisting}[style=mystyle, caption={Part illustration of modulation used in \ourwork (in Python-like codes).}, label={code:mod}]
# self: the instance of the AdaLN method
# self.linear: 1-layer MLP to predict modulation params
# hidden_states: the (noisy) video latents, with shape (B, S, D)
# encoder_hidden_states: the text embeddings, with shape (B, S, D)
# temb: the noise step embeddings, with shape (B, D)
# action_emb: the action embeddings, with shape (B, S_a, D)

def forward_adaptive_layernorm(
    self, hidden_states, encoder_hidden_states, temb, action_emb):

    # Vision Expert AdaLN (timestep + action)
    embedding_dim = hidden_states.shape[-1]
    shift, scale, gate = torch.nn.functional.linear(
        self.silu(temb[:, None, :] + action_emb),
        self.linear.weight[: 3 * embedding_dim],
        self.linear.bias[: 3 * embedding_dim],
    ).chunk(3, dim=-1)

    # Text Expert AdaLN (only timestep)
    enc_shift, enc_scale, enc_gate = torch.nn.functional.linear(
        self.silu(temb),
        self.linear.weight[3 * embedding_dim :],
        self.linear.bias[3 * embedding_dim :],
    ).chunk(3, dim=-1)

    # Modulate Vision Hidden States
    num_patches = hidden_states.size(1) // action_emb.size(1)
    scale = scale.repeat_interleave(repeats=num_patches, dim=1)
    shift = shift.repeat_interleave(repeats=num_patches, dim=1)
    hidden_states = self.norm(hidden_states) * (1 + scale) + shift

    # Modulate Text Hidden States
    encoder_hidden_states = self.norm(encoder_hidden_states) * \
            (1 + enc_scale)[:, None, :] + enc_shift[:, None, :]
    ...
\end{lstlisting}
\end{minipage}
\vspace{-10pt}
\end{figure*}

\subsection{Video Generation Details (Section~\ref{sec:vgm})}
\label{sec:supp_model_arch}

\subsubsection{Model Details}

\mypar{Hyperparameters.} As mentioned in Sec.~\ref{sec:vgm}, we use the CogVideoX-2B\footnote{\href{https://huggingface.co/zai-org/CogVideoX-2b}{https://huggingface.co/zai-org/CogVideoX-2b} (including VAE, T5 and transformers)}~\cite{yang2024cogvideox} as our pretrained backbone, which is a compromise between training from scratch and using the larger pretrained model (\textit{e.g.}, CogVideoX-5B as TesserAct~\cite{zhen2025tesseract}). And we have already shown its better performance than training from scratch (see Tab.~\ref{tab:abla_cond}) and strong generalization ability in the experiments (see Fig.~\ref{fig:diverse}). We list the main hyperparameters of the model architecture in Tab.~\ref{tab:supp_model}.

\mypar{Modulations.} CogVideoX~\cite{yang2024cogvideox} adopts an Expert Adaptive LayerNorm design, where the diffusion timestep $t$ is fed into a modulation module that produces parameters for both the Vision and Text Expert AdaLNs to modulate their respective hidden states (vision and text). Since our model is initialized from the pretrained CogVideoX, we retain this architecture to preserve its generation capability. To incorporate 3D action control, we repurpose the Vision Expert AdaLN—originally designed to modulate vision hidden states—to apply modulation from action inputs, while keeping the Text Expert AdaLN unchanged (see Listing~\ref{code:mod}).

\mypar{Multiple Visual Conditions.} To fuse multiple visual conditioning inputs (depth and semantics), we first concatenate the multiple condition latents along the channel dimension, then repeat the input noise latents and add them to the condition latents. After that, we reduce the channels back to the same as the noise latents. As illustrated in Eq.~\ref{eq:cond_in}, where $z_{\text{in}}$ represents the input noise latents.
\begin{equation}
\label{eq:cond_in}
    z_{\text{in}}=\mathrm{MLP}(z_{\text{in}}+\mathrm{Concat}([c_1,c_2,\dots])) + z_{\text{in}}
\end{equation}

\begin{table*}[!ht]
\centering
\begin{minipage}[t]{0.58\textwidth}
\renewcommand{\arraystretch}{0.95}
\centering
\caption{Hyperparameters of data preprocessing for training and evaluations, where $\Delta f_1$ represents the sample interval of frames within video samples, $\Delta f_2$ represents the sample interval among video samples of split data (train, val).}
\label{tab:supp_data}
\setlength{\tabcolsep}{3.5pt}
\begin{tabular}{c c c c c c c}
\toprule[1pt]
 & frames & raw size & sample size & latent size & $\Delta f_{1}$ & $\Delta f_{2}$ \\
\midrule[0.5pt]
BridgeV2~\cite{walke2023bridgedata} & 16 & 480$\times$640 & 320$\times$480 & 40$\times$60 & 1 & 4,\;16 \\
DROID~\cite{khazatsky2024droid} & 24 & 180$\times$256 & 256$\times$384 & 32$\times$40 & 3 & 16,\;72 \\
RT-1~\cite{brohan2022rt1} & 16 & 256$\times$320 & 320$\times$480 & 40$\times$60 & 2 & 6,\;16 \\
\bottomrule[1pt]
\end{tabular}
\end{minipage}
\hfill
\begin{minipage}[t]{0.38\textwidth}
\renewcommand{\arraystretch}{0.95}
\centering
\caption{Distributions of multiview data of BridgeData V2~\cite{walke2023bridgedata}.}
\label{tab:data_multiview}
\setlength{\tabcolsep}{6pt}
\begin{tabular}{c c c}
\toprule[1pt]
& samples & proportion(\%) \\
\midrule[0.5pt]
n\_view=1 & 89901 & 60.79 \\
n\_view=2 & 0 & 0.00 \\
n\_view=3 & 57978 & 39.21 \\
total & 147879 & 100.00 \\
\bottomrule[1pt]
\end{tabular}
\end{minipage}
\end{table*}

\begin{figure*}
    \centering
    \includegraphics[width=\textwidth]{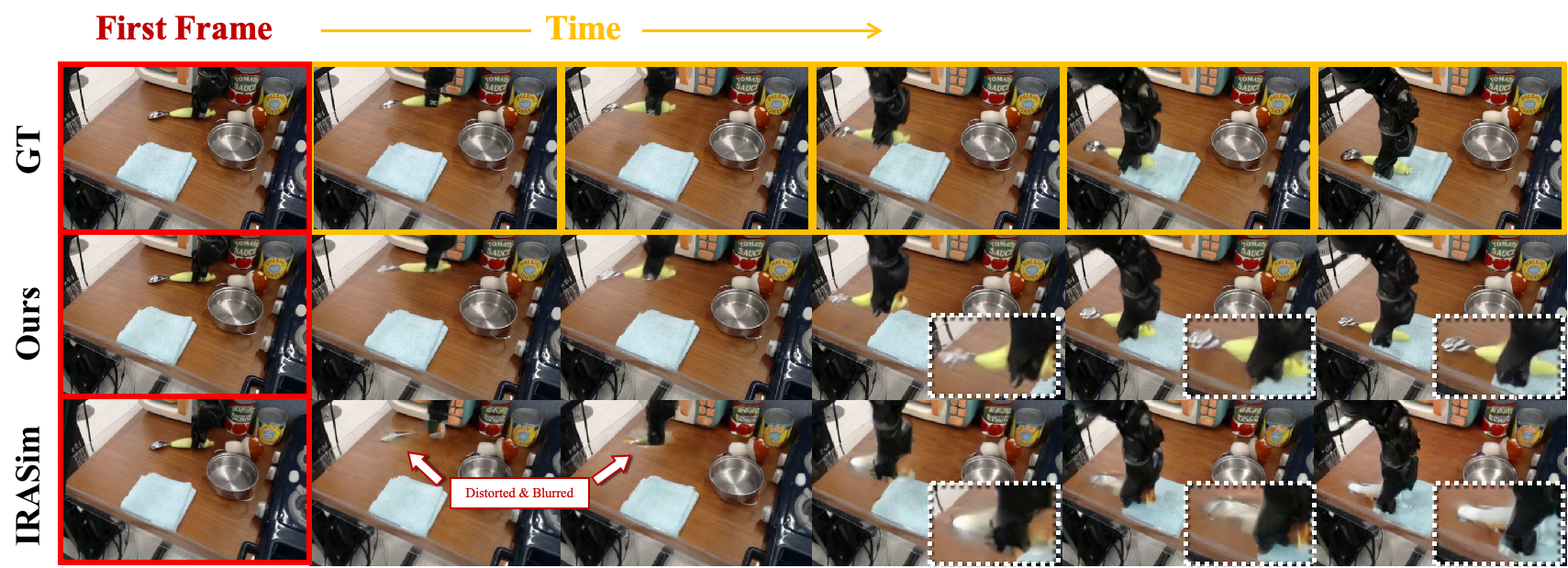}
    \caption{Qualitative Results of \boldourwork with full conditions. \textcolor{red}{Red boxes} denote the first frame input of the video generation; \textcolor{orange}{Orange boxes} denote the ground-truth of the subsequence frames.}
    \label{fig:recon2}
\end{figure*}

\mypar{Positional Encoding.} We use the 3D sincos positional encodings in DiT blocks, following the original CogVideoX-2B. In our multiview videos generation model, similar to the temporal 3D positional encoding applied on singleview videos, we apply another spatial 3D positional encoding which is added to the multiview images for each single frame (as Eq.~\ref{eq:pe}). It will enable our model to learn to operate each view accordingly since the order and the number of the input views during training is constantly randomized.
\begin{equation}
\label{eq:pe}
\begin{aligned}
\mathrm{PE}(t,x,y) &= \mathrm{PE}_t(t) \oplus \mathrm{PE}_s(x,y) &&\rightarrow\quad \text{Frame Attn.} \\
\mathrm{PE}(v,x,y) &= \mathrm{PE}_v(v) \oplus \mathrm{PE}_s(x,y) &&\rightarrow\quad \text{View Attn.}
\end{aligned}
\end{equation}

\mypar{3D VAE.} The unique design of 3D VAE of CogVideoX requires the input videos to have a length of $8N+1$ where $N\leq6$. To accommodate this requirement, we append an additional single frame to the end of each sequence, which merely serves as a placeholder (\textit{e.g.}, if we train and test the sequence length of 16, then we exactly input a 17-frame sequence into the model). It will ensure the model encodes (decodes) the videos (latents) correctly. Simply, we directly discard the last frame after the VAE decoding during evaluation. As for the action sequence, to ensure the latent-frame-level alignment, we also append a subsequent action to the last frame. And to be compatible with the chunk-level injection (as introduced in Sec.~\ref{sec:vgm}) where the chunk size is exactly equal to the temporal compression ratio of 3D VAE, we again pad another ($\mathrm{chunk\_size-1}$) zeros to the last frame. Hence, the last $\mathrm{chunk\_size}$ actions actually serve as the placeholders in our model.

\begin{figure*}
    \centering
    \includegraphics[width=\textwidth]{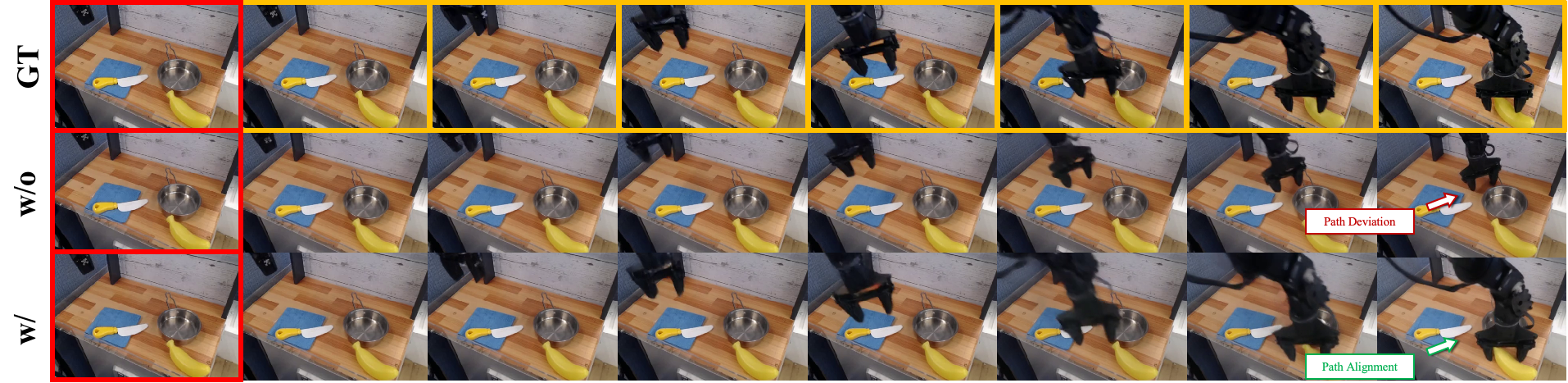}
    \caption{Ablation Results of \textbf{Depth Condition Map.} Without any physical controls, the robot gripper fails to act accurately aligned with the 3D action instructions, due to the accumulation of errors. While ours performs correctly, along with the entire sequence.}
    \label{fig:abla_depth}
\end{figure*}

\begin{figure*}
    \centering
    \includegraphics[width=\textwidth]{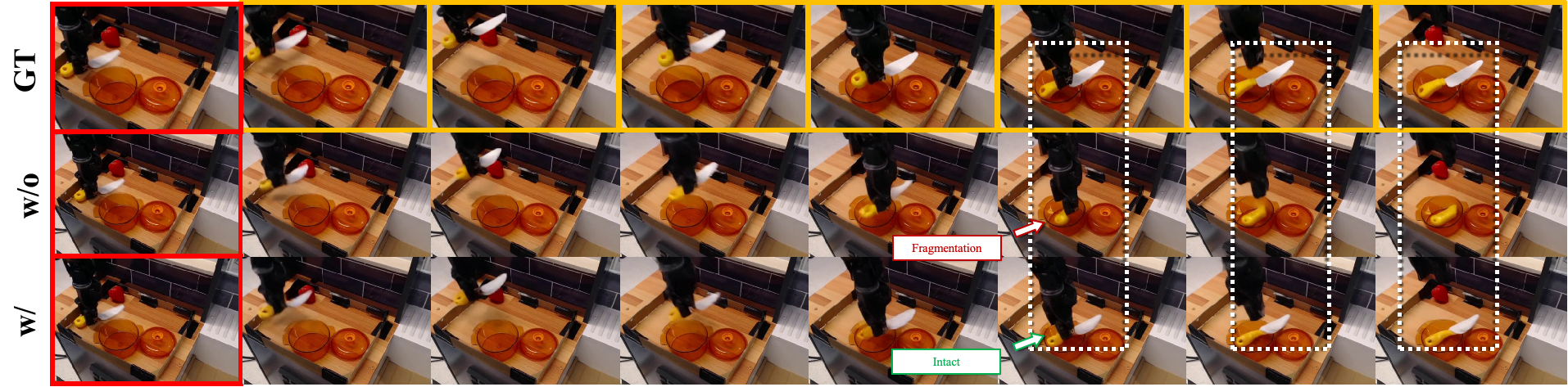}
    \caption{Ablation Results of \textbf{Semantics Condition Map.} Without the guidance of our rendered semantics maps, the model fails to accurately predict the shape deformation of the knife during its motion, whereas ours produce outputs that align well with the real-world appearance.}
    \label{fig:abla_sem}
\vspace{-15pt}
\end{figure*}

\subsubsection{Training Details}

\mypar{Data Process.} During training, we sample sequences of frames by first randomly selecting a video and then uniformly sampling a segment of a specified length and size. Given the various raw resolutions of videos in different datasets (as introduced in Sec.~\ref{sec:supp_data}), we process them into a similar resolution setting for stable training. Moreover, the datasets are recorded at different frequencies (\textit{e.g.}, the robot gripper in BridgeV2 data moves much faster than that in DROID data). To maintain consistency, we sample the sequences at varied step sizes. Taking into account all these factors (resolutions, sampling frequencies), we also set different sequence lengths to ensure that each sequence can ideally capture a complete operation, while controlling the total number of visual tokens of each sample to be processed by the model. Take the BridgeV2 singleview training as an example, each individual sample will result in a total $\lceil(16+1)/4\rceil\times(40 / 2\times60/2)=3000$ tokens. We list all the details mentioned above in Tab.~\ref{tab:supp_data}.
Note that the number of total frames of each individual episode varies significantly across the datasets (\textit{e.g.}, 20$\sim$50 for BridgeV2 while 50$\sim$4000 for DROID). We then take different sample intervals, \textit{i.e.}, the interval between the neighboring sequences within the same episode, for training and evaluation.

\mypar{Multiview Generations.} In our training of the multiview videos generation model, we control the proportion of samples with varying numbers of views in the training data to ensure both effective and robust learning. Specifically, taking the BridgeData V2~\cite{walke2023bridgedata} dataset as an example, the full set of training samples generated through sampling contains a total of 147,879 samples. Among these, 60.79\% consist of only a single view, while 39.21\% have three views. To balance the data, we randomly subsample from the singleview group to reduce its proportion to around 40\%. During training, we randomly sample the number of views from the sample data. Specifically, we have the probability of 0.5 to sample a 2-view sequence and another 0.5 to have a 3-view sequence, when the current sample has 3 views. Furthermore, to facilitate the training of multiview generation model, we initialize the weights of the multiview module in \ourwork-MV (shown in Fig.~\ref{fig:mv_pipeline}) directly through copying from the singleview module.

\subsection{Policy Learning Details (Section~\ref{sec:exp_policy})}
\label{sec:supp_model_policy}

\subsubsection{Data Augmentation}

As demonstrated in Sec.~\ref{sec:exp_policy}, with augmented manipulation data powered by \ourwork, the policy learning of various vision-language-action (VLA) models can be significantly improved, suggesting a promising direction for leveraging \textit{generative world model} to enhance policy learning \textit{with low costs}.
Recent concurrent works following this pardigm have also demonstrate remarkable success, including DreamGen~\cite{jang2025dreamgen}, RoboBrain-X0~\footnote{\href{https://github.com/FlagOpen/RoboBrain-X0}{https://github.com/FlagOpen/RoboBrain-X0}}, Gigabrain-0~\cite{team2025gigabrain}, Emma~\cite{dong2025emma}, MimicDreamer~\cite{li2025mimicdreamer}, EmbodiedDreamer~\cite{wang2025embodiedreamer}, EgoDemoGen~\cite{xu2025egodemogen}.
In our experiments, we primarily focus on augment the existing BridgeData V2~\cite{walke2023bridgedata}, enhancing the visual diversity in a real-to-real manner.

\begin{figure*}[t]
\centering
\begin{minipage}{0.92\textwidth}
\begin{lstlisting}[style=mystyle, caption={Text instruction used in Qwen2.5-32B-Instruct for video captioning.}, label={code:qwen2.5_caption}]
Output only one sentence that describes what the robot arm or gripper is doing. The sentence must strictly start with a verb, not with 'The robot arm' or any subject. Do not use 'is', 'I am', or 'The robot arm'. Only output the instruction in imperative form.
\end{lstlisting}
\end{minipage}
\vspace{-10pt}
\end{figure*}

\begin{figure*}[t]
\centering
\begin{minipage}{0.92\textwidth}
\begin{lstlisting}[style=mystyle, caption={Part of text instruction used in Qwen2.5-32B-Instruct for video evaluation.}, label={code:qwen2.5}]
You are a strict, reliable, and accountable video quality evaluator specialized in robot-arm manipulation videos. Follow the rules exactly. If you break them, you will be penalized.

ROLE:
- You are an impartial human-like evaluator.
- Behave like a careful human reviewer: inspect frames, identify problems, and assign fair scores.
- Do NOT hallucinate or guess unseen details.

INPUT:
- A video of robot manipulation (assume frames and timestamps are accessible).

TASK:
1. Inspect the entire video. Pay attention to serious defects like geometry collapse, object deformation, or impossible motion.
2. Score the video on each criterion (1-5 scale).
3. For each criterion, only output:
- numeric score (1..5)
- confidence (0.00-1.00)
4. Do NOT output justification for every criterion (to save tokens).
5. Instead, provide ONE short `"summary"` (only 1 sentence) describing the main issues.
6. Compute `"final_score"` as weighted average (weights below).

CRITERIA (score each 1..5):
A. clarity - sharpness, focus, absence of blur, no ghosting_artifacts.
B. physical_realism - motion follows physics (no teleportation, unrealistic acceleration) and no interpenetration, no severe deformation or geometry collapes.
C. overall_plausibility - temporal/spatial consistency, lighting stability, no sudden jumps.
\end{lstlisting}
\end{minipage}
\vspace{-10pt}
\end{figure*}

\mypar{Data Generation.}
Given the full conditions, We directly use the model in Sec.~\ref{sec:exp_video_gen} for singleview video generation.
Similar to Sec.~\ref{sec:supp_s2r}, we employ the x-flux model with ControlNet to generate the initial frames based on the conditions from the dataset, yielding three manipulation videos with difference appearance.
Then, we use the large vision-language-model (VLM), Qwen2.5-32B-Instruct\footnote{\href{https://huggingface.co/Qwen/Qwen2.5-32B-Instruct}{https://huggingface.co/Qwen/Qwen2.5-32B-Instruct}}~\cite{qwen2.5} to caption all generated videos, using the designed prompt shown in Listing~\ref{code:qwen2.5_caption}.
Ultimately we obtain additional $\sim$40K synthesized videos based on samples randomly drawn from the dataset.

\mypar{Data Cleaning.}
While augmenting the dataset for greatly improved visual diversity, some generated samples still exhibit poor quality (\textit{e.g.}, unrealistic deformations, blur, or implausible manipulations) that can hinder the policy training.
We further employ VLM for efficient data filtering.
Specifically, we use Qwen2.5-32B-Instruct~\cite{qwen2.5} to exhaustively score all generated videos given carefully-designed prompts, partly shown in Listing~\ref{code:qwen2.5}.
Each video is evaluated along three aspects---visual clarity, physical realism, and overall plausibility---to remove those containing blurry appearances, ghosting artifacts, physically implausible motions (\textit{e.g.}, severe deformation or geometry collapse), or temporal inconsistencies.
In our experiments, approximately 10\% of the data were filtered out.

\begin{figure*}
    \centering
    \includegraphics[width=\textwidth]{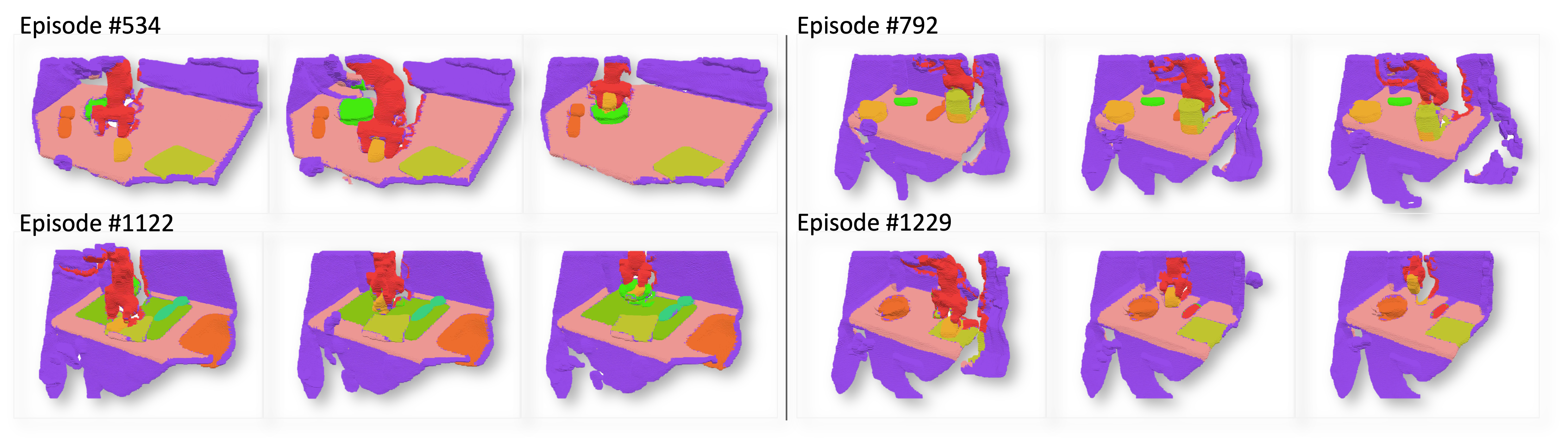}
    \vspace{-20pt}
    \caption{Additional examples of 4D semantic occupancy data (on BridgeData V2~\cite{walke2023bridgedata}) used in \ourwork.}
    \label{fig:more_occ}
    \vspace{-5pt}
\end{figure*}

\begin{figure*}
    \centering
    \includegraphics[width=\textwidth]{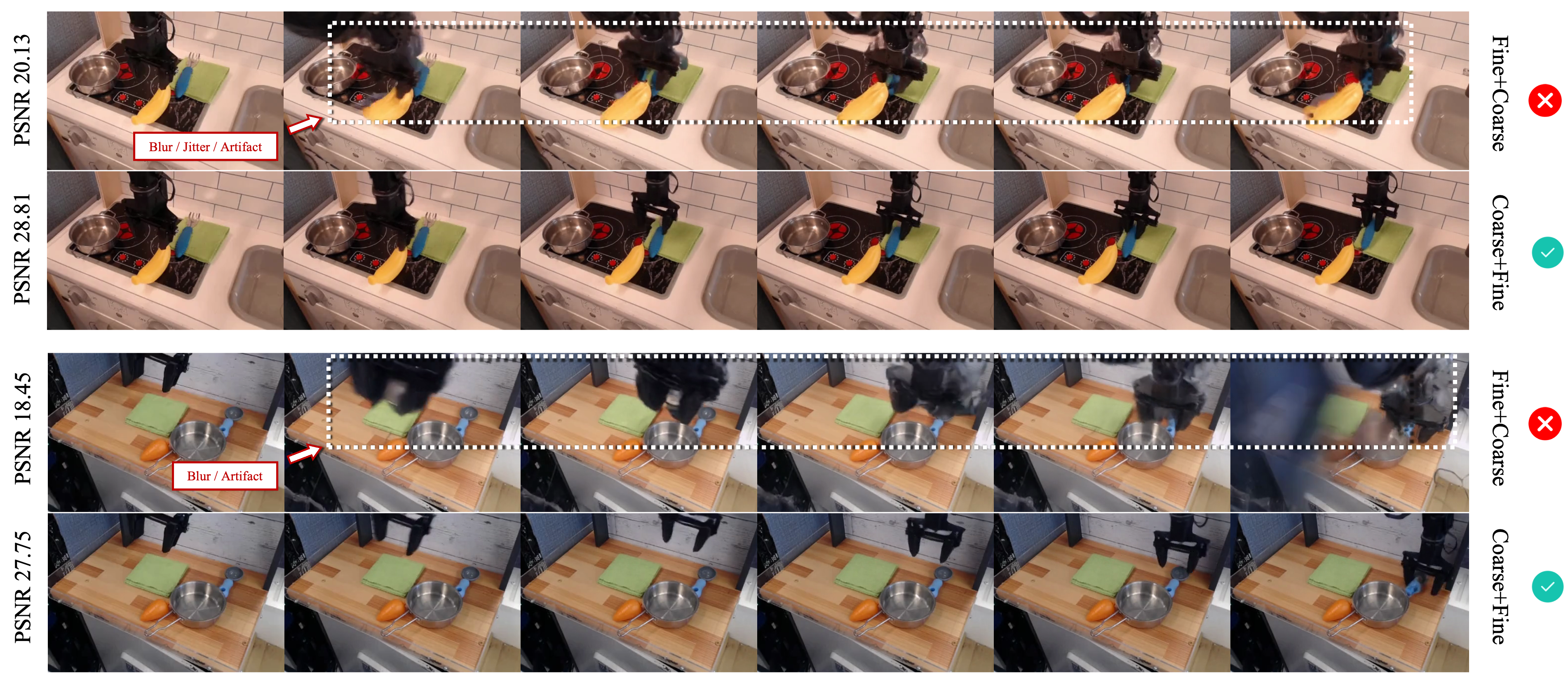}
    \vspace{-15pt}
    \caption{Qualitative comparison of zero-shot conditional video generation under different occupancy conditioning sources (refer to Tab.~\ref{tab:abla_zero_shot}). The label ``A+B'' denotes training on A and evaluating on B. ``Fine'' indicates condition maps that are accurately aligned with the ground truth at the pixel level, while ``Coarse'' refers to those derived from occupancy fields. The PSNR values are calculated against the ground truths.}
    \label{fig:abla_fine_coarse}
    \vspace{-10pt}
\end{figure*}

\subsubsection{VLA Post-Finetuning}

Current mainstream vision-action-language (VLA) models~\cite{qu2025spatialvla,kim2024openvla} usually follow a two-stage training paradigm: 1) Pretraining on large-scale cross-embodiment manipulation data (millions of samples, \textit{e.g.}, Open-X-Embodiment~\cite{open_x_embodiment_rt_x_2023}) to acquire general action-planning capabilities; and 2) Finetuning on in-domain datasets to enhance task performance (\textit{e.g.}, BridgeData V2~\cite{walke2023bridgedata} on SimplerEnv-WidowX~\cite{li2024evaluating}, LIBERO Data~\cite{liu2023libero} on the LIBERO Benchmark).
Notably, RoboVLM~\cite{liu2025towards} systematically explored different VLA training strategies, including: a) \textit{In-domain Finetuning}, directly train VLA on in-domain datasets; b) \textit{OXE Pretrain}, pre-train the VLA on OXE dataset; and c) \textit{Post-finetuning}, train the OXE-pretrained VLA on in-domain datasets---a two-stage strategy that yields superior performance.
In our experiments, we also adopt the approach c), post-finetuning after cross-embodiment pre-training, to evaluate the data augmentations.

For RoboVLM~\cite{liu2025towards}, we use oxe-pretrained-robovlm~\footnote{\href{https://huggingface.co/robovlms/RoboVLMs/blob/main/checkpoints/kosmos_ph_oxe-pretrain.pt}{https://huggingface.co/robovlms/RoboVLMs}} as the pretrained model, which is adapted from \texttt{kosmos-2}~\footnote{\href{https://huggingface.co/docs/transformers/model_doc/kosmos-2}{https://huggingface.co/docs/transformers/model\_doc/kosmos-2}}, and follow the open-sourced scripts for full finetuning.
For SpatialVLA~\cite{qu2025spatialvla}, we use oxe-pretrained-spatial~\footnote{\href{https://huggingface.co/IPEC-COMMUNITY/spatialvla-4b-224-pt}{https://huggingface.co/IPEC-COMMUNITY/spatialvla-4b-224-pt}} as the pretrained model, which is adapted from \texttt{paligemma}~\footnote{\href{https://huggingface.co/blog/paligemma}{https://huggingface.co/blog/paligemma}} and follow the open-sourced scripts for LoRA finetuning.

\begin{figure*}
    \centering
    \includegraphics[width=\textwidth]{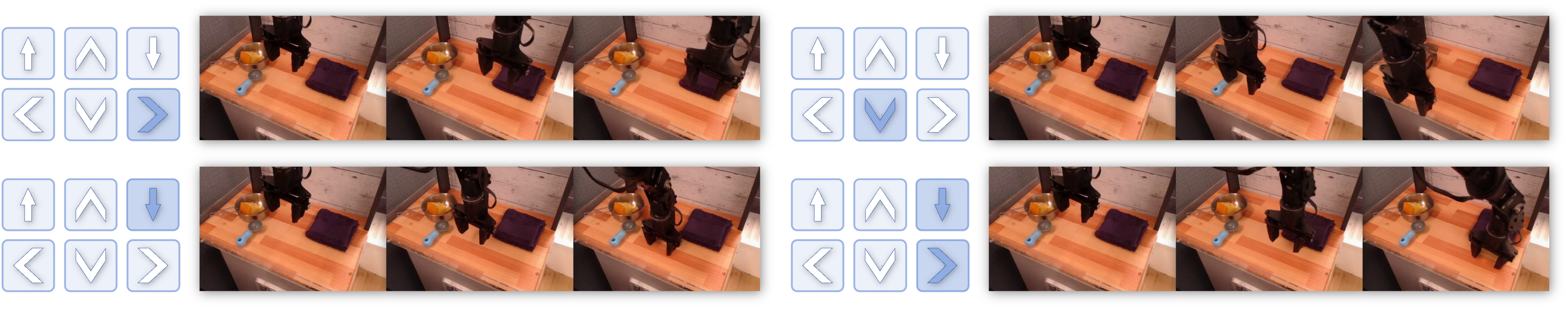}
    \vspace{-15pt}
    \caption{Qualitative results of action-conditioned video generation, where varying input actions enable precise control over the gripper. Better to zoom in.}
    \label{fig:action_control}
    \vspace{-10pt}
\end{figure*}

\begin{figure}
    \centering
    \includegraphics[width=\linewidth]{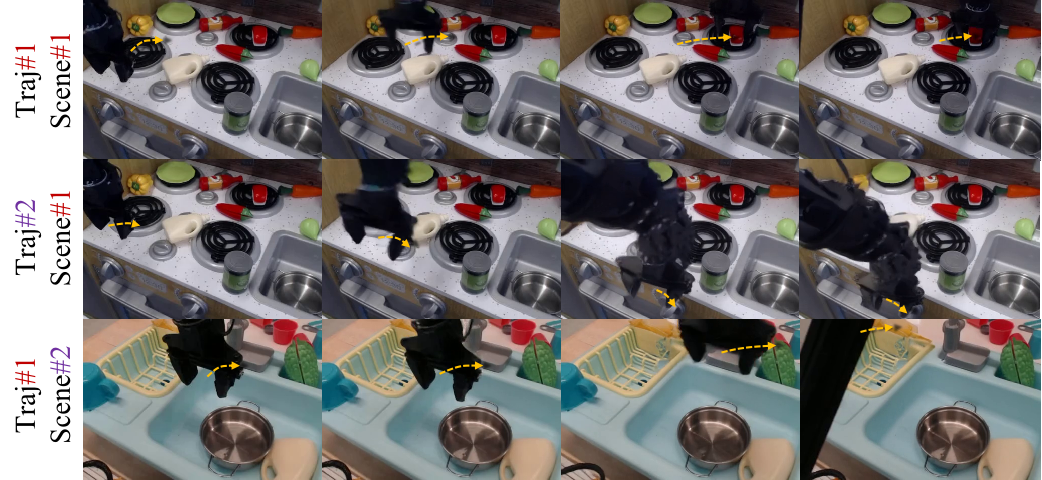}
    \captionsetup{font=footnotesize}
    \caption{\textbf{Appearance \& Trajectory} Adaptation Results. Zoom in for better observations.}
    \vspace{-20pt}
    \label{fig:in_domain}
\end{figure}

\subsection{Evaluation Details}

For \textit{conditional video generation}, we evaluate our model across four common metrics: Peak Signal-to-Noise Ratio (PSNR)~\cite{huynh2008psnr}, Structural Similarity Index Measure (SSIM)~\cite{wang2004ssim}, Fr\'echet Inception Distance (FID)~\cite{heusel2017gans} and Fr\'echet Video Distance (FVD)~\cite{unterthiner2018fvd}.
All of our evaluations involve the $\sim$2.6K of generated samples.
For \textit{visual planning}, we strictly follow the settings of VP$^2$~\cite{tian2023control} benchmark to calculate the success rate.
For \textit{policy learning}, we also strictly follow the instructions of SIMPLER~\cite{li2024evaluating} to conduct the evaluation process.

\subsection{Computation Resources}

We implement \ourwork in PyTorch, using the \texttt{diffusers}\footnote{\href{https://github.com/huggingface/diffusers}{https://github.com/huggingface/diffusers} under Apache License} and \texttt{transformers}\footnote{\href{https://github.com/huggingface/transformers}{https://github.com/huggingface/transformers} under Apache License} libraries. Our models are trained and evaluated on an $\text{8}\times \text{H100}$ cluster. Each experiment utilizes 8 GPUs in parallel, with 16 data loader workers per device. Since we use the similar volume of tokens and size of models in calculation and size of training samples across different datasets, each single 30K-gradient-step training costs around 35 hours ($\sim$11.7 GPU days) and evaluating $\sim$3K samples will cost nearly 2 hours (also parallel in 8 GPUs). Dataset curation particularly cost much disk space, \textit{e.g.}, all generated data for BridgeData V2~\cite{walke2023bridgedata} in our experiments occupies about 8TB of disk space.

\section{Additional Results}
\label{sec:supp_exp}

In this section, we present additional experimental details and results.
For \textit{conditional video generation}, we include extended comparisons and ablations on control signals, as well as more examples demonstrating the generalization ability and multiview video generations of \ourwork.
For \textit{policy learning}, we provide the details of data augmentations and qualitative results illustrating policy evaluation augmented by \ourwork.

\subsection{Curated Occupancy Data}
\label{sec:more_occ}
Additional examples of 4D occupancy data are shown in Fig.\ref{fig:more_occ}, complementary to Fig.\ref{fig:data}.

\subsection{Controllable Video Generation (Section~\ref{sec:exp_video_gen})}
\label{sec:supp_exp_gen}

\mypar{Baselines.} We compare our results with recent open-sourced works. \textbf{IRASim}~\cite{zhu2024irasim} is a video diffusion model employing DiT architecture with action modulation, which outperforms both VDM~\cite{ho2022vdm} and LVDM~\cite{he2022lvdm}. \textbf{HMA}~\cite{wang2025hma} models video dynamics via a masked autoregressive transformer tailored for real-world action sequences. \textbf{AVID} designs a plug-in adapter that can inject action controls to pretrained video generation models.

\begin{figure*}
    \centering
    \includegraphics[width=\textwidth]{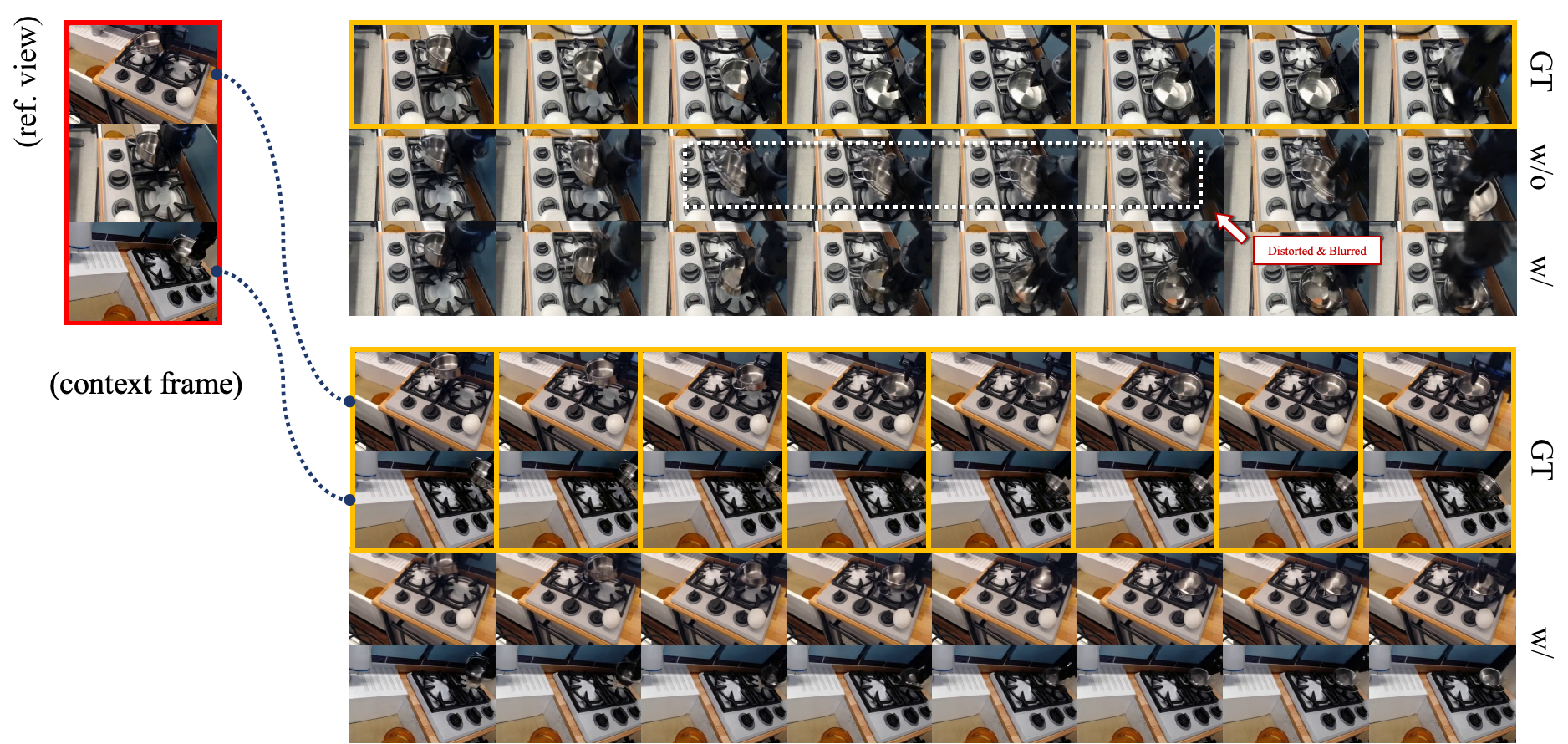}
    \caption{Qualitative Comparison Results of \textbf{Multiview Videos Generation}. With our from-reference-view rendered visual conditionings, generated videos under side views achieve better geometric consistency under other side views. Better to zoom in.}
    \label{fig:mv_cond}
    \vspace{-10pt}
\end{figure*}

\begin{figure}[t]
\centering
\begin{minipage}[t]{0.49\columnwidth}
    \includegraphics[width=\linewidth]{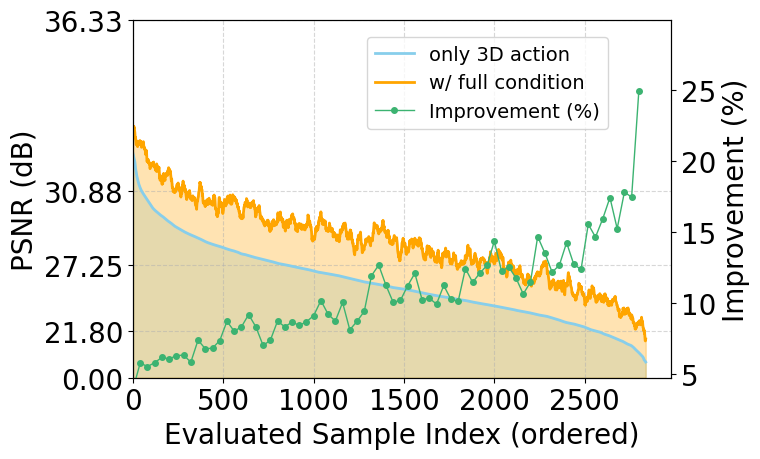}
\end{minipage}
\hfill
\begin{minipage}[t]{0.49\columnwidth}
    \includegraphics[width=\linewidth]{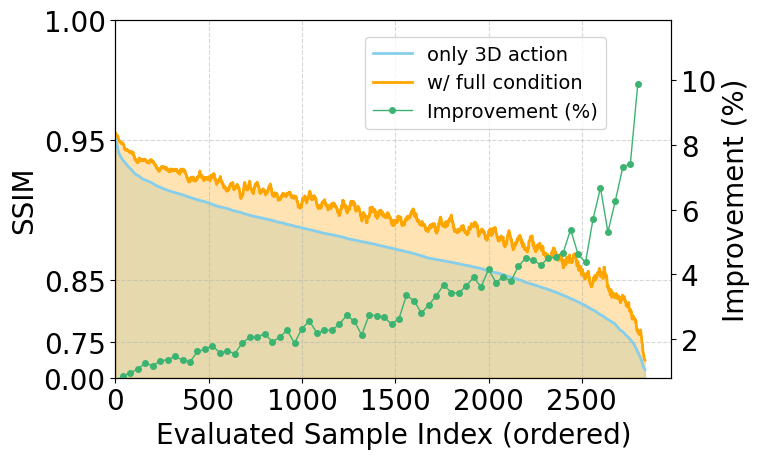}
\end{minipage}
\captionsetup{font=footnotesize}
\caption{Improvement curves of PSNR (left) and SSIM (right) metrics across ordered evaluation samples from BridgeData V2~\cite{walke2023bridgedata}.}
\label{fig:metric_curve}
\vspace{-20pt}
\end{figure}

\mypar{More Comparison with Baselines.}
As shown in Fig.~\ref{fig:recon2}, we provide another comparison between IRASim~\cite{zhu2024irasim} and \ourwork.
As highlighted by the red indicators and white boxes, the baseline fails to accurately reconstruct the physical appearance of the object manipulated by the robot gripper during motion.
Such dynamics are crucial for downstream applications such as policy and imitation learning.
In contrast, \ourwork demonstrates more faithful and consistent generation of the interaction process.

\begin{figure}
    \centering
    \includegraphics[width=\linewidth]{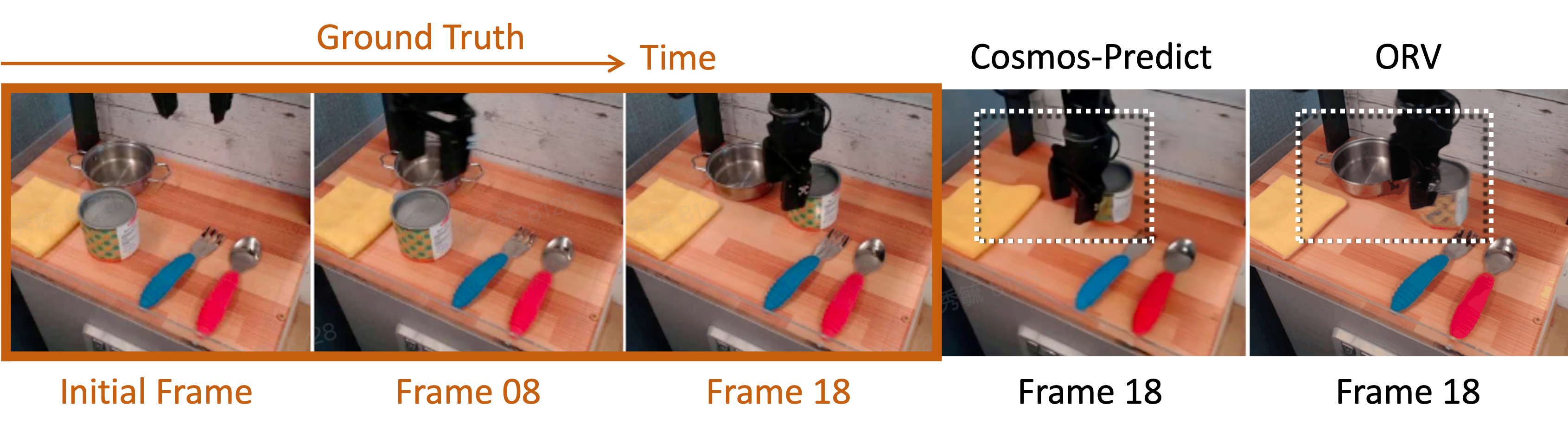}
    \captionsetup{font=footnotesize}
    \vspace{-20pt}
    \caption{Comparisons between Cosmos~\cite{agarwal2025cosmos} and \ourwork generated rollouts..}
    \vspace{-10pt}
    \label{fig:cosmos_compare}
\end{figure}

\mypar{Cosmos Comparison.}
We also evaluate our approach against Cosmos~\cite{agarwal2025cosmos}, a recent strong foundation model (by NVIDIA).
As shown in Fig.~\ref{fig:cosmos_compare}, rollouts generated by Cosmos are susceptible to history distortion, leading to noticeable temporal inconsistencies and structural degradation over time.
Conversely, ORV effectively mitigates this issue and demonstrates remarkable temporal robustness.
By leveraging 4D occupancy-centric conditioning, our model maintains strict spatio-temporal coherence and high visual fidelity throughout the entire sequence, ensuring physically plausible environmental dynamics.

\mypar{Effect of Control Signals.}
We present quantitative results in Tab.~\ref{tab:abla_cond} to demonstrate the improvements brought by incorporating physical control signals, and visualize the effects in Fig.~\ref{fig:abla_depth}.
As shown, without depth guidance, the robot gripper fails to accurately follow the 3D action instructions—an expected result since 2D pixels are inherently insensitive to depth variations.
In contrast, with our rendered depth conditions, this limitation is effectively mitigated.
Fig.~\ref{fig:abla_sem} further provides qualitative comparisons with and without semantic condition maps, showing clear improvements when semantic priors are introduced.

We further aggregate evaluation scores across all samples to analyze the effect of incorporating occupancy-based guidance.
Using the BridgeData V2~\cite{walke2023bridgedata} as an example, Fig.~\ref{fig:metric_curve} illustrates the sample-wise improvements in PSNR and SSIM after applying the full conditioning.
Specifically, we first sort all samples according to their scores under the base model---\textit{e.g.}, using only 3D action conditions (\textcolor{SkyBlue}{blue curve})—and then plot the corresponding scores obtained with the full conditioning (\textcolor{orange}{orange curve}) following the same order.
The green curve further indicates the per-sample relative improvement (\%).

\mypar{Robustness of Occupancy Representations.}
The qualitative results that demonstrate the robustness of our introduced occupancy representations are shown in Fig.~\ref{fig:abla_fine_coarse}, complementary to Tab.~\ref{tab:abla_zero_shot}.
We can clearly observe that when \ourwork is trained on ``Fine’’ condition maps, a hard constraint emerges: only condition inputs with similar granularity can retain the optimal performance of the model, which substantially limits its applicability across diverse scenarios.

\subsubsection{Generalizations}

Despite being adapted from a pretrained CogVideoX model via SFT, \ourwork demonstrates strong generalization capability, delivering robust performance across diverse robotic manipulation scenarios.
Beyond the quantitative results in Sec.~\ref{tab:policy_bridge}, Fig.~\ref{fig:in_domain} further illustrates \ourwork’s video generations under diverse appearances and arbitrary action modifications, exhibiting both precise controllability and consistent generalization.
Additionally, Fig.~\ref{fig:action_control} showcase the manipulation video generation where the robot gripper is controlled by random external action inputs. 
However, as \ourwork does not utilize textual prompts and instead relies solely on visual cues to infer the states of robot arms and grippers, it is not yet capable of executing semantically meaningful tasks.

\subsubsection{Multiview Videos Generation}

Maintaining consistency across different views is crucial for multiview video generation. Although the model may possess the ability to infer view orientations from the observed frame (referred to as the context frame) and to predict how 3D motion control translates into 2D pixel variations across views, this capability is inherently limited. Therefore, we provide multiview conditioning signals consistently rendered from 3D geometric representations to enhance 2D pixel predictions, as described in Sec.~\ref{sec:mv} and Sec.~\ref{sec:supp_mv}.

Fig.~\ref{fig:mv_cond} compares a 3-view video generation with and without the additional conditioning maps. In this example, although the 3D occupancy is constructed solely from the anchor view due to data limitations---resulting in lower quality compared to a complete 3D geometry---the conditioning maps rendered from the other two side views still improve the overall generation quality. As highlighted by the white regions, during the robot gripper’s motion while holding a metal bowl, the bowl exhibits severe deformation in the current view, even though this issue is entirely absent in the anchor view. This discrepancy mainly arises from two factors: (1) the current view differs significantly from the anchor view, and (2) the object undergoes relatively large motion. With the additional guidance from 3D geometry, these issues can be effectively mitigated.

\subsubsection{Real-World Validation (\textit{Zero-shot})}
\label{sec:supp_real_world}
\begin{figure}
    \centering
    \includegraphics[width=0.9\linewidth]{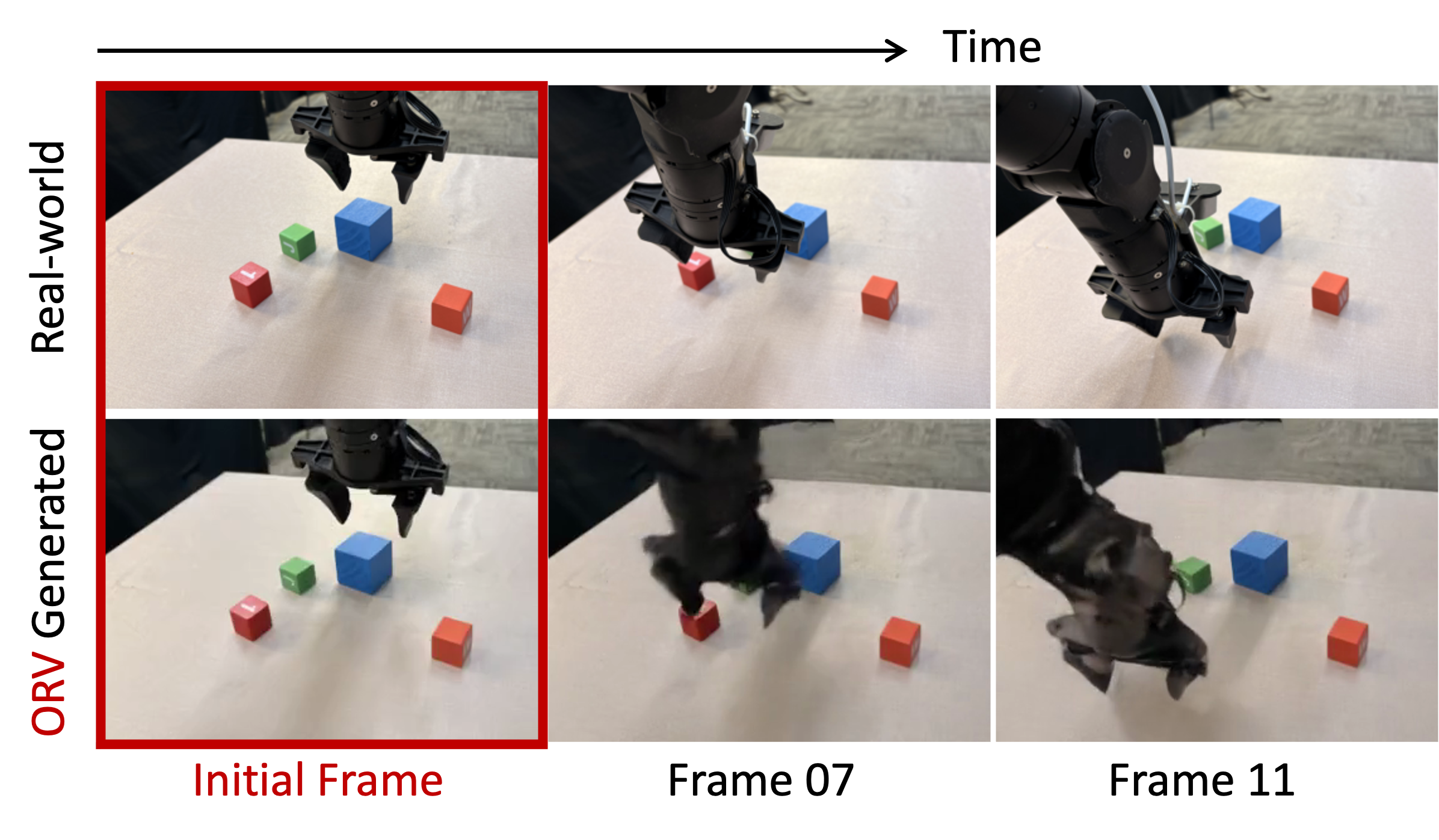}
    \captionsetup{font=footnotesize}
    \vspace{-10pt}
    \caption{Real-world rollout and generated rollout (\textit{zero-shot}).}
    \vspace{-15pt}
    \label{fig:real_world}
\end{figure}

To further validate the zero-shot generalization capability of ORV in practical scenarios, we conducted an evaluation using a physical AgileX Piper robotic arm. Given its morphological similarity to the WidowX platform used in our model , we captured a real-world manipulation rollout and fed the identical action sequence into our pre-trained model without any domain-specific fine-tuning. As illustrated in Fig.~\ref{fig:real_world}, the ORV-generated rollout demonstrates high visual fidelity and strong physical alignment with the real-world ground truth, confirming the model's robust ability to synthesize faithful environmental dynamics directly from unseen physical inputs.

\subsubsection{Additional Qualitative Results}
\label{sec:supp_gallery}

\begin{figure*}
    \centering
    \includegraphics[width=.9\textwidth]{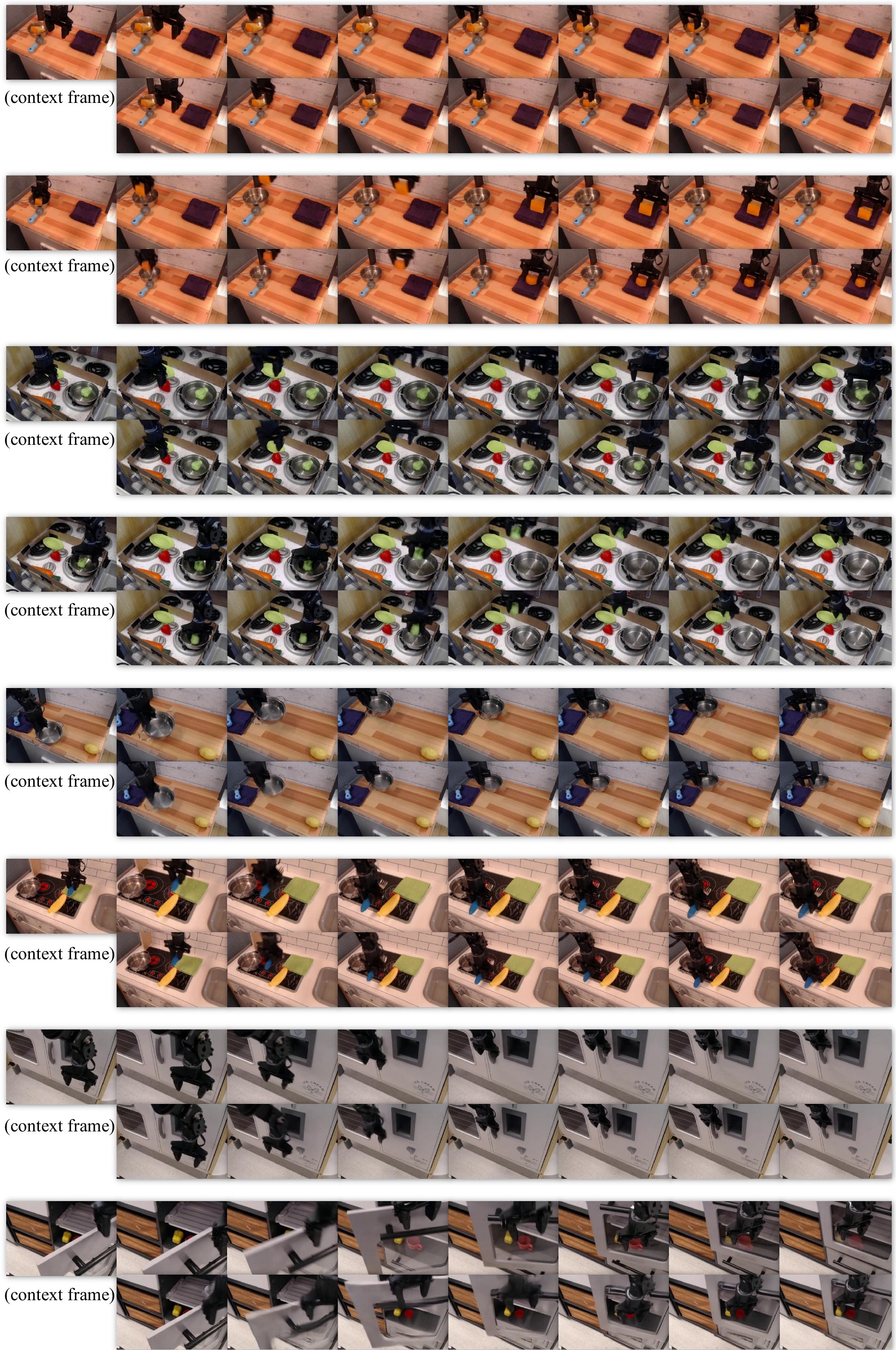}
    \caption{Additional Qualitative Results of \ourwork \#1.}
    \label{fig:gallery_1}
\end{figure*}

\begin{figure*}
    \centering
    \includegraphics[width=.9\textwidth]{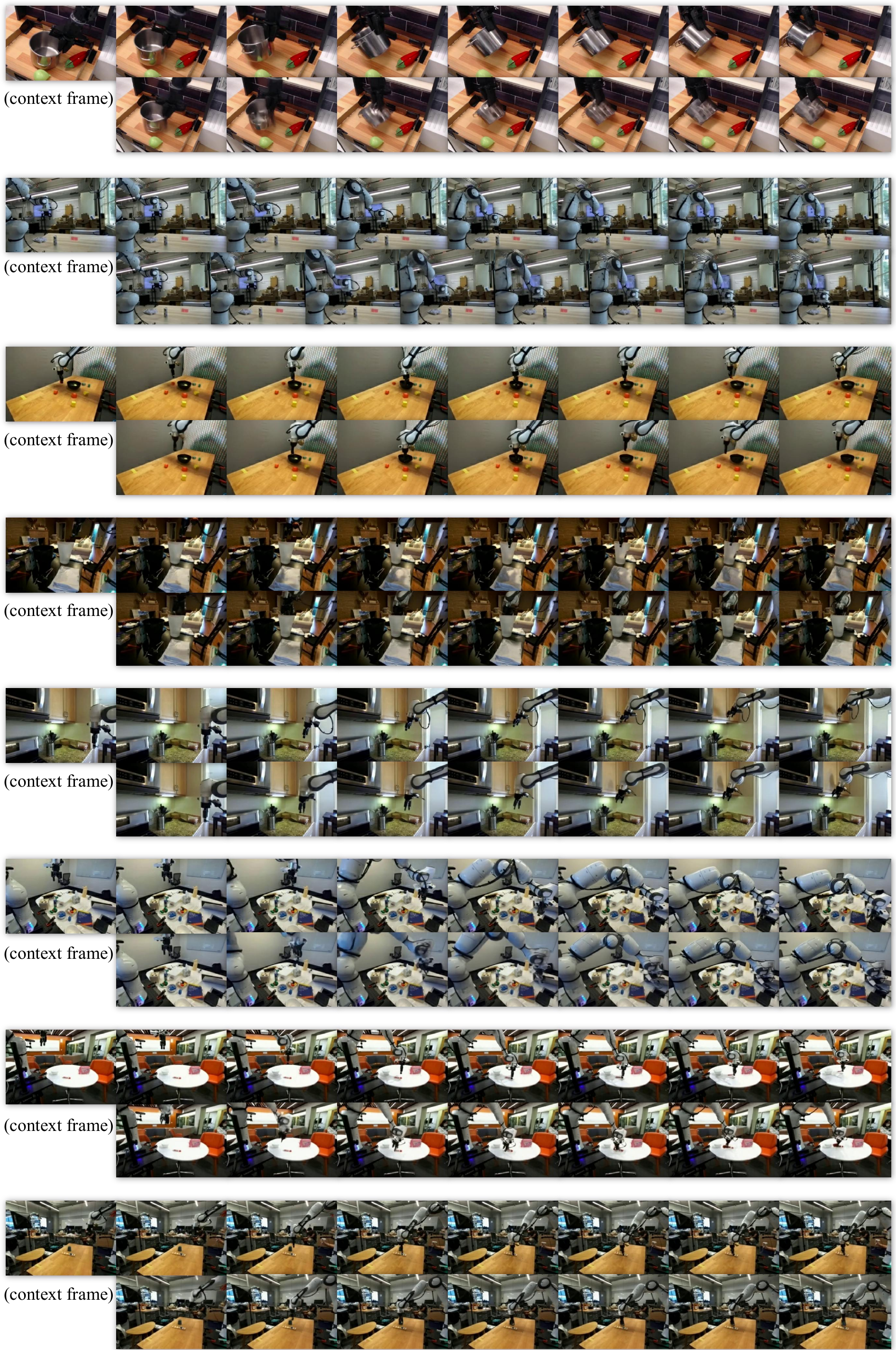}
    \caption{Additional Qualitative Results of \ourwork \#2.}
    \label{fig:gallery_2}
\end{figure*}

\begin{figure*}
    \centering
    \includegraphics[width=.9\textwidth]{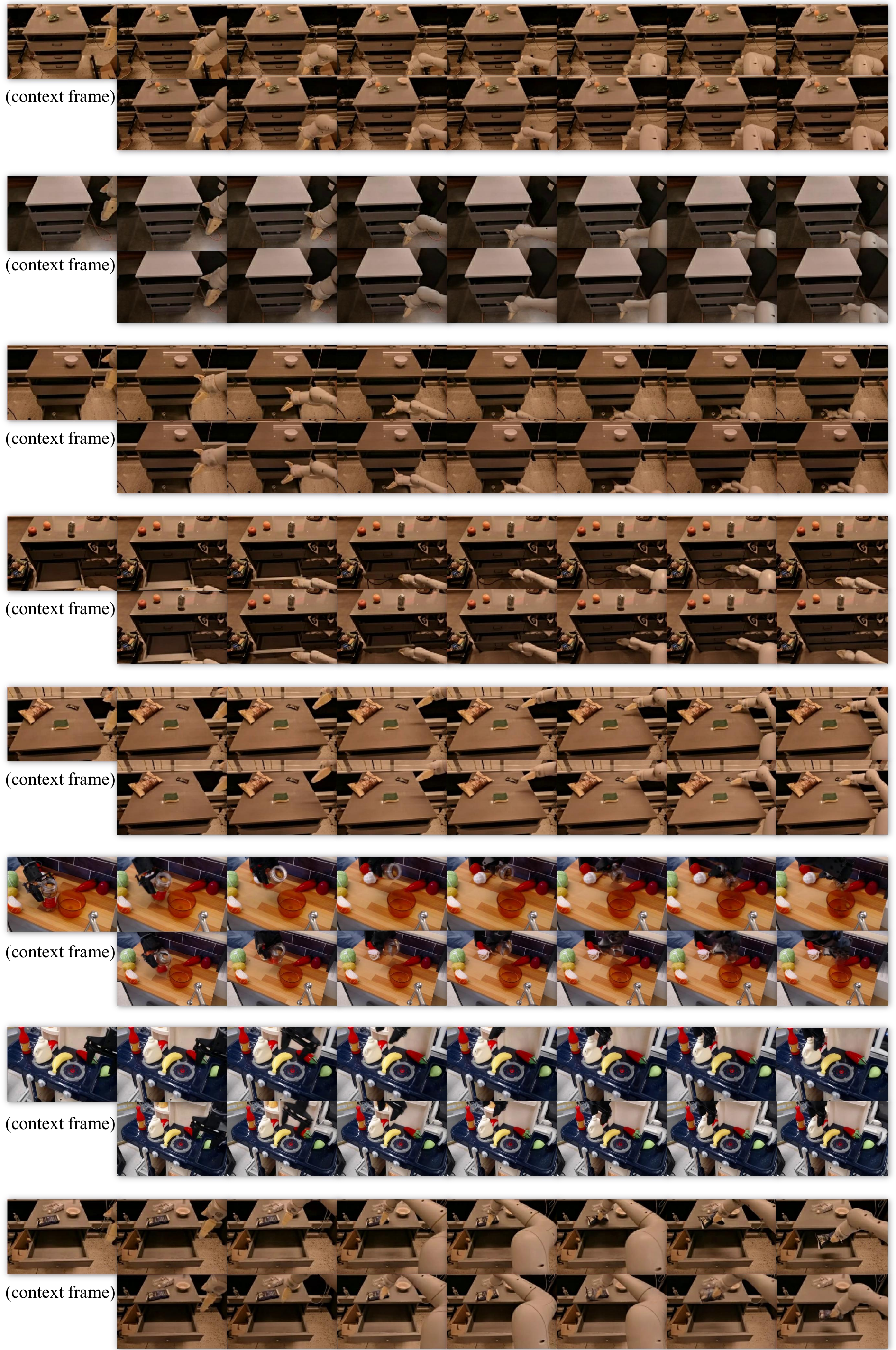}
    \caption{Additional Qualitative Results of \ourwork \#3.}
    \label{fig:gallery_3}
\end{figure*}

We provide more \textbf{uncurated} singleview examples generated by \ourwork, as shown in Fig.~\ref{fig:gallery_1},~\ref{fig:gallery_2},~\ref{fig:gallery_3}. For each episode, we present their ground-truths in the top row and our results in the bottom row, respectively. For a better view and other more examples, please refer to our webpage.

\subsection{Policy Learning (Section~\ref{sec:exp_policy})}
\label{sec:supp_policy_learning}

\begin{figure*}
    \centering
    \includegraphics[width=\textwidth]{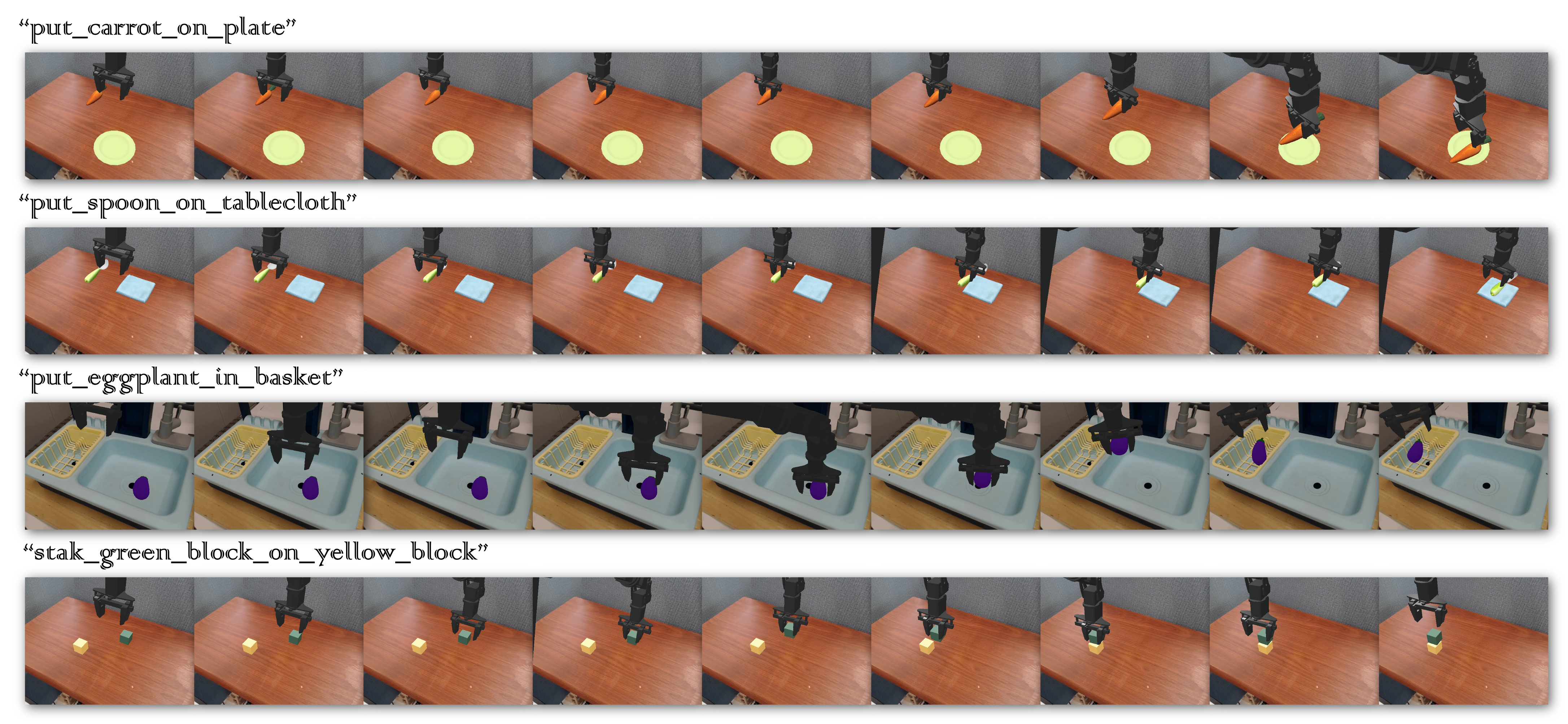}
    \vspace{-15pt}
    \caption{Successful examples of policy execution on four SimplerEnv-WidowX~\cite{li2024evaluating} tasks using our fine-tuned SpatialVLA~\cite{qu2025spatialvla} model.}
    \label{fig:policy_video}
    \vspace{-12pt}
\end{figure*}

Fig.~\ref{fig:policy_video} shows four successful execution examples of finetuned SpatialVLA~\cite{qu2025spatialvla} with augmented data, on SimplerEnv-WidowX~\cite{li2024evaluating} Benchmark.

\section{Discussions}
\label{sec:supp_discuss}

In this section, we provide more insightful explanations of \ourwork model or more in-depth discussion of extended related works, covering a broader range of aspects concerning generative models for robotics.

\subsection{Occupancy-centric Framework}
\label{sec:discuss_occ}

\mypar{Action and Visual Priors.}
As described in Sec.\ref{sec:formulation}, most recent controllable video generation approaches for robot manipulation follow an action-to-video paradigm\cite{zhu2024irasim,wang2025hma,guo2025ctrl}, where 3D action values are recorded either in simulation environments or from real-world robots. Some works, such as Im2Flow2Act~\cite{xu2024flow}, instead employ pixel-level 2D flows as intermediate motion signals, while others explore trajectory- or action-conditioned generation beyond robotics, \textit{e.g.}, Tora~\cite{zhang2024tora} for human-drawn trajectory control.
Although encoding 3D actions has shown promising results, the abstract nature of these values limits the model’s ability to infer complex future object states—particularly for motions orthogonal to the image plane or involving rotations (see Fig.\ref{fig:abla_depth}). In contrast, visual cues such as 2D pixel flows provide more precise and stable motion guidance but remain insufficient to describe the entire scene. Furthermore, visual priors used in Cosmos-Transfer\cite{alhaija2025cosmos} and RoboTransfer~\cite{liu2025robotransfer} require pixel-perfect alignment with ground-truth depth or segmentation maps, which is often infeasible.
To address these limitations, we combine \textit{high-level, hard} action priors with \textit{low-level, soft} visual priors rendered from occupancy fields. This hybrid design ensures that the generated videos follow action instructions while allowing flexible, coarser visual conditioning, thus mitigating the constraints of previous approaches.

\mypar{Occupancy Representation.}
Occupancy fields offer multiple advantages beyond providing robust representations of noisy or parametric scenes, as discussed in Sec.\ref{sec:formulation}.
Their coordinate-based formulation enables efficient online forecasting of robot manipulation scenes—directly predicting future states of the environment in the occupancy space.
This paradigm has demonstrated remarkable success in autonomous driving\cite{wei2023surroundocc,huang2023tri,wang2023openoccupancy,li2024uniscene,tian2024occ3d,chen2025trackocc,wang2025diffusion,han2025unicalib,wang2025unifying}, where occupancy representation has become a preferred choice over 3D points, bounding boxes, or meshes.
Numerous recent works, including OccSora~\cite{wang2024occsora}, Occllama~\cite{wei2024occllama}, OccFormer~\cite{zhang2023occformer}, OccGen~\cite{wang2024occgen}, and OccWorld~\cite{zheng2023occworld}, have achieved high-quality 3D occupancy generation and forecasting, highlighting a promising direction toward extending occupancy forecasting to robotic manipulation.
Although robotics presents greater challenges due to more complex scene dynamics, achieving online 3D occupancy forecasting would further reduce the reliance on physical simulators that allow policy networks to generate the dynamics, facilitating the acquisition of occupancy priors for \ourwork.

\mypar{Occupancy Data Curation.}
As introduced in Sec.\ref{sec:data}, we curate a 4D occupancy dataset for robotic manipulation by leveraging multiple foundation models within the data curation pipeline, including MonST3R\cite{zhang2024monst3r}, NKSR~\cite{huang2023nksr}, VLMs~\cite{bai2023qwen}, and SAM2~\cite{ravi2024sam}.
In our experiments, such scene reconstruction models demonstrate strong reliability on large-scale robotic datasets, effectively capturing fine-grained object and gripper motions.
Moreover, with the incorporation of our \textit{soft} visual priors, the reliance on precise dynamic modeling during reconstruction is further reduced, making the overall data generation pipeline both robust and scalable.

\mypar{Non-interactive Generative Model.}
Different from the recent work iVideoGPT~\cite{wu2024ivideogpt}, which highlights its interactive capability, our \ourwork mainly focuses on non-interactive generation.
An interactive generation framework (typically, an auto-regressive model) exhibits causality during the forward pass, enabling arbitrary interactions with the external physical world.
For instance, compared to iVideoGPT, VideoGPT~\cite{yan2021videogpt} only accepts the entire future action sequence at the start of prediction, preventing an agent from interactively adjusting its actions based on predicted observations.
However, our \ourwork model adopts the architecture of a non-causal diffusion model, where the action and occupancy priors are fully acquired before generation. Consequently, it does not require any interaction with the external world during generation.

\subsection{World Model for Robot Manipulation}

A world model is an internal abstraction that captures the physical, spatial, and causal dynamics of an environment.
It encodes multimodal inputs (\textit{e.g.}, images, text, actions, audio) into latent representations and predicts future states through internal reasoning and simulation. 
In robot manipulation, it models the interaction between sensory observations and actions.

\mypar{World Model for High-fidelity Simulation.}
\ourwork serves as a generative world model that simulates diverse real-world environments.
To ensure the simulated dynamics closely resemble real-world physical behaviors, \ourwork achieves superior performance compared to recent approaches such as IRASim~\cite{zhu2024irasim} and HMA~\cite{wang2025hma}.
We primarily evaluate the effectiveness of the world model in simulation by comparing the visual quality of its predictions against ground-truth observations using standard metrics (\textit{e.g.}, PSNR, FVD).
Concurrent efforts~\cite{wang2025sampo,guo2025ctrl,fu2025learningvideogenerationrobotic,liu2025geometry} have also been exploring high-fidelity and physically accurate video simulations.

\mypar{World Model for Efficient Data Synthesis.}
Sec.\ref{sec:supp_policy_learning} and Sec.\ref{sec:exp_policy} have introduced how \ourwork benefits policy learning through data synthesis.
While some recent works~\cite{zhu2024irasim,dong2025emma,guo2025ctrl} share similar ideas, they differ from ours.
IRASim~\cite{zhu2024irasim} deploys a pretrained policy model in a simulator to generate additional rollouts—both successful and failed ones—for training world models.
RoboTransfer~\cite{liu2025robotransfer} trains a synthesis model with decoupled geometry and appearance conditions, where the conditions are derived from real-world data.
Ctrl-World~\cite{guo2025ctrl} generates synthetic post-training data by either rephrasing task instructions or resetting the robot arm to a new initial state for additional trajectories, which are then used for policy training.
As we can see, most of these approaches require a data preparation stage to utilize the world model as a data generator.
While we mainly demonstrate and validate the \textit{visual} transfer capability of \ourwork in our experiments, we argue that \ourwork can also generate manipulation videos with diverse trajectories for each task.
In such cases, we would similarly deploy a pretrained policy model (\textit{e.g.}, RoboVLM~\cite{liu2025towards}, SpatialVLA~\cite{qu2025spatialvla}, $\pi_0$~\cite{black2410pi0} etc.) in a physical simulator and perform simulation-to-real generation as discussed in Sec.~\ref{sec:sim2real}.

\mypar{World Model for Reproducible Policy Evaluation.}
Taking the world model as a simulator that accurately mimics the real world, some recent works~\cite{zhu2024irasim,guo2025ctrl} explore developing reproducible policy evaluation within the world model itself.
In this way, the world model can be used to evaluate upstream policy models, just as in the real world, while significantly reducing computational and physical resources.
Although \ourwork could potentially support such functionality, we clarify that it is beyond the scope of this paper.

\subsection{Manipulation Data Augmentation}
\label{sec:supp_data_aug}
Data augmentation for policy learning typically involves appearance, trajectory/action, and viewpoint.
We further discuss how \ourwork handles these forms of augmentation.

\mypar{Appearance Augmentation.}
As demonstrated in Sec.~\ref{sec:exp_policy} and Sec.~\ref{sec:supp_model_policy}, we mainly augment existing manipulation data through appearance randomization for policy learning, by leveraging additional image generator.
Appearance augmentation improves policy robustness by exposing the policy to a broader distribution of textures, lighting conditions, and background variations, thereby reducing overfitting to the visual biases of the original dataset.

\mypar{Trajectory Augmentation.}
Trajectory augmentation exposes the policy to diverse state–action patterns, improving robustness under distribution shifts.
To achieve this, one typically leverages a physical simulator to execute policies and collect varied rollouts.
For instance, Ctrl-World~\cite{guo2025ctrl} increases rollout diversity by (1) rephrasing task instructions or (2) resetting the robot arm to new initial states.
We leave such trajectory-level augmentation to future work, as it is not the primary focus of this work.

\mypar{Viewpoint Augmentation.}
For each manipulation task, current policy models typically rely on single-view demonstrations, as high-quality multiview manipulation data is often unavailable.
Acquiring multiview data requires accurate 3D geometry of the scene.
Given that \ourwork leverages 3D occupancy representations, it remains the potential to support viewpoint augmentation; in this way, an additional model is needed to generate multiview initial frames from these geometric priors, which represents a promising direction for future work.

\begin{figure}
    \centering
    \includegraphics[width=\linewidth]{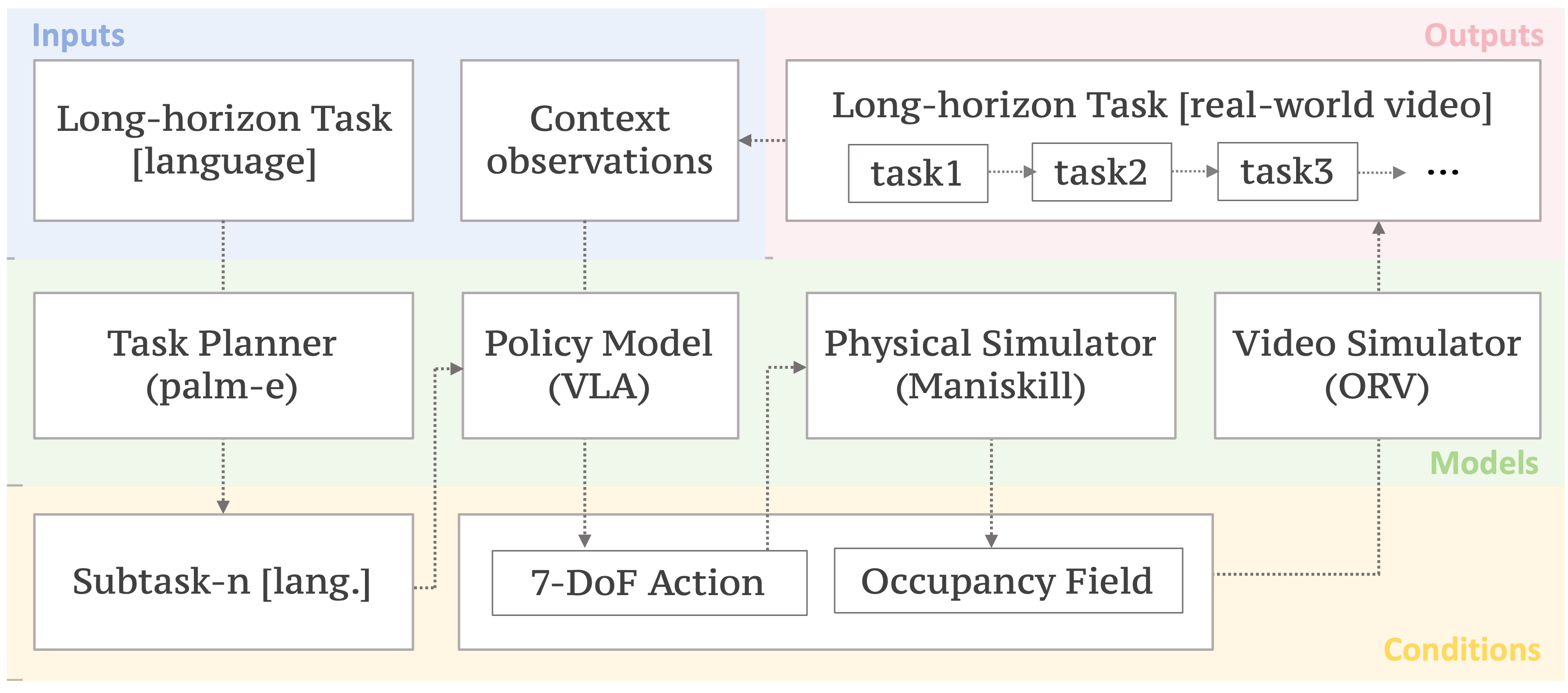}
    \captionsetup{font=footnotesize}
    \caption{System of long-horizon manipulation data synthesis.}
    \vspace{-10pt}
    \label{fig:long}
\end{figure}

\subsection{Limitations and Future Directions.}
Despite the promising results achieved by \ourwork, the task remains inherently challenging with numerous open issues, and our approach has certain limitations.
In this section, we elaborate on these limitations and suggest possible avenues for future research.

\begin{itemize}
    \item \textit{Integrating online 4D occupancy generation or forecasting.} Currently, \ourwork relies on complete 4D occupancy data as input, which to some extent limits its applicability in real-world scenarios. As discussed in Sec.~\ref{sec:discuss_occ}, the coordinate-based formulation of occupancy representations, combined with the success of online occupancy generation frameworks in autonomous driving, suggests that incorporating such an online 4D occupancy generation or forecasting module into \ourwork is both feasible and promising. This integration would enable real-time perception and significantly enhance the practicality of our work.
    \item \textit{Incorporating more comprehensive action representation of the robot arm.} Although our 3D occupancy provides a comprehensive geometric representation of all objects in the scene, the 3D action signal in our framework only encodes the 7-DoF end-effector pose of the robotic arm. Such a description is insufficient for manipulators with more complex articulated points, such as the Google robot used in the DROID~\cite{khazatsky2024droid} dataset—where rich joint-level dynamics are essential for accurately modeling the motion. Incorporating detailed motion descriptions for all joints would therefore yield a more faithful and fine-grained representation of the arm’s trajectory. And recent work VAP~\cite{wang2025precise} provides an alternative.
    \item \textit{Adding multiview initial frames generations to \ourwork-MV.} Specifically, \ourwork-MV requires the first-frame observations from multiple camera views. By leveraging geometric constraints from the 3D occupancy and the robotic arm pose observed in these initial frames, \ourwork-MV can generate view-consistent videos. In future work, we plan to extend this framework to synthesize multiview first-frame images directly from a singleview input—\textit{i.e.}, enabling consistent multiview video generation from only one camera view. Such an enhancement would greatly improve the scalability and real-world usability of \ourwork-MV.
    \item \textit{Towards long-horizon robot manipulation planning and generation.} Long-horizon manipulation data are substantially more valuable for policy training~\cite{RoboBrain2.0TechnicalReport,mees2022calvin,han2025robocerebra} yet much harder to collect than short-horizon video data. We believe that extending \ourwork into a long-horizon manipulation data planning and generation framework is feasible and promising (as illustrated in Fig.~\ref{fig:long}). Such an extension would enable the synthesis of temporally coherent long-horizon manipulation data, thereby facilitating more challenging policy learning and improving generalization across complex tasks.
\end{itemize}

\subsection{Social Impact}

This work advances controllable robot video generation with broad applications in robotics simulation, education, virtual reality, and creative media. Acknowledging its dual-use risks, such as potential misuse for misinformation or privacy violations, we conduct all research under a responsible AI framework using ethically sourced, public datasets for academic purposes only. We advocate incorporating safeguards like provenance tracking and synthetic content detection to ensure generative technologies benefit society while minimizing harm.

\section{License}
\begin{itemize}
    \item All datasets used for video generation (BridgeData V2~\cite{walke2023bridgedata}, DROID~\cite{khazatsky2024droid}, RT-1~\cite{brohan2022rt1}) are maintained under CC-BY-4.0 License;
    \item Robusuite~\cite{robosuite2020}: MIT License;
    \item Robodesk~\cite{kannan2021robodesk}: Apache License 2.0;
    \item Qwen-VL-Chat~\cite{bai2023qwen}: released under Qwen-VL License Agreement\footnote{\href{https://github.com/QwenLM/Qwen-VL/blob/master/LICENSE.txt}{https://github.com/QwenLM/Qwen-VL/blob/master/LICENSE}};
    \item Qwen2.5-32B-Instruct~\cite{qwen2.5}: Apache License 2.0;
    \item sentence-transformers/all-MiniLM-L6-v2~\cite{wang2020minilmv2}: Apache License 2.0;
    \item CogVideoX-2B~\cite{yang2024cogvideox}: Apache License 2.0;
    \item MonST3R~\cite{zhang2024monst3r}: MIT License;
    \item VGGT~\cite{wang2025vggt}: released under VGGT License~\footnote{\href{https://github.com/facebookresearch/vggt/blob/main/LICENSE.txt}{https://github.com/facebookresearch/vggt/blob/main/LICENSE}};
    \item NKSR~\cite{huang2023nksr}: Apache License 2.0;
    \item RAFT~\cite{teed2020raft}: BSD 3-Clause License;
    \item Grounding DINO~\cite{liu2024grounding}: Apache License 2.0;
    \item SegmentAnything2~\cite{ravi2024sam}: Apache License 2.0;
    \item ManiSkill~\cite{gu2023maniskill2}: code and rigid-body environment components are released under Apache License 2.0; Assets are licensed under CC BY-NC 3.0;
    \item SIMPLER Benchmark~\cite{li2024evaluating}: MIT License;
    \item X-FLUX: all pretrained models are under FLUX.1 [dev] Non-Commercial License\footnote{\href{https://github.com/black-forest-labs/flux/blob/main/model_licenses/LICENSE-FLUX1-dev}{https://github.com/black-forest-labs/flux/blob/main/model\_licenses/LICENSE-FLUX1-dev}}; codes are under Apache License 2.0;
\end{itemize}